%% file: arxiv.tex
\documentclass[10pt,twocolumn,letterpaper]{article}

\usepackage{iccv}
\usepackage{times}
\usepackage{epsfig}
\usepackage{graphicx}
\usepackage{amsmath}
\usepackage{amssymb}

\usepackage{tablefootnote}
\usepackage{multirow}
\usepackage{xcolor} 
\usepackage{verbatim}
\usepackage{url}
\usepackage{amssymb}
\usepackage{mathrsfs}
\usepackage{caption,tabularx,booktabs}
\usepackage{adjustbox}
\usepackage{diagbox}
\usepackage{balance}
\usepackage{mathrsfs}
\usepackage{pifont}
\newcommand{\cmark}{\ding{51}}%
\newcommand{\xmark}{\ding{55}}%

\iccvfinalcopy 

\def\iccvPaperID{****} 

\ificcvfinal\fi

\begin{document}

\title{Voxel-based Network for Shape Completion by Leveraging Edge Generation}

\author{Xiaogang Wang\quad\quad Marcelo H Ang Jr\quad\quad Gim Hee Lee\\
National University of Singapore\\
{\tt\small xiaogangw@u.nus.edu\quad \{mpeangh,gimhee.lee\}@nus.edu.sg}}

\maketitle
\ificcvfinal\thispagestyle{empty}\fi

\begin{abstract}
Deep learning technique has yielded significant improvements in point cloud completion with the aim of completing missing object shapes from partial inputs. However, most existing methods fail to recover realistic structures due to over-smoothing of fine-grained details. In this paper, we develop a voxel-based network for point cloud completion by leveraging edge generation (VE-PCN). We first embed point clouds into regular voxel grids, and then generate complete objects with the help of the hallucinated shape edges. This decoupled architecture together with a multi-scale grid feature learning is able to generate more realistic on-surface details. We evaluate our model on the publicly available completion datasets and show that it outperforms existing state-of-the-art approaches quantitatively and qualitatively. Our source code is available at \url{https://github.com/xiaogangw/VE-PCN}.
\end{abstract}

\section{Introduction}
3D shape completion is a fundamental problem in computer vision and robotic perception.
The aim is to reconstruct complete object topologies from sparse and incomplete observations, \emph{e.g.} raw data collected by RGB-D or LiDAR sensors. 
Since incompleteness and irregularity of input point clouds impose difficulties on down-stream tasks such as 3D object classification~\cite{qi2017pointnet,qi2017pointnet++,wang2019dynamic,li2018so}, segmentation~\cite{qi2017pointnet,qi2017pointnet++,li2018pointcnn} and detection~\cite{chen2017multi,qi2018frustum,yang2018pixor,zhou2018voxelnet}, several point cloud completion methods~\cite{yuan2018pcn,topnet2019,Wang_2020_CVPR,liu2020morphing,wen2020point,sarmad2019rl} are proposed to improve the quality of the point clouds. 
Although existing works have achieved impressive results, they are limited to
low-fidelity outputs.
In this work, we focus on generating high-quality 3D objects from occluded inputs.

\begin{figure}
    \centering
  \includegraphics[width=\linewidth]{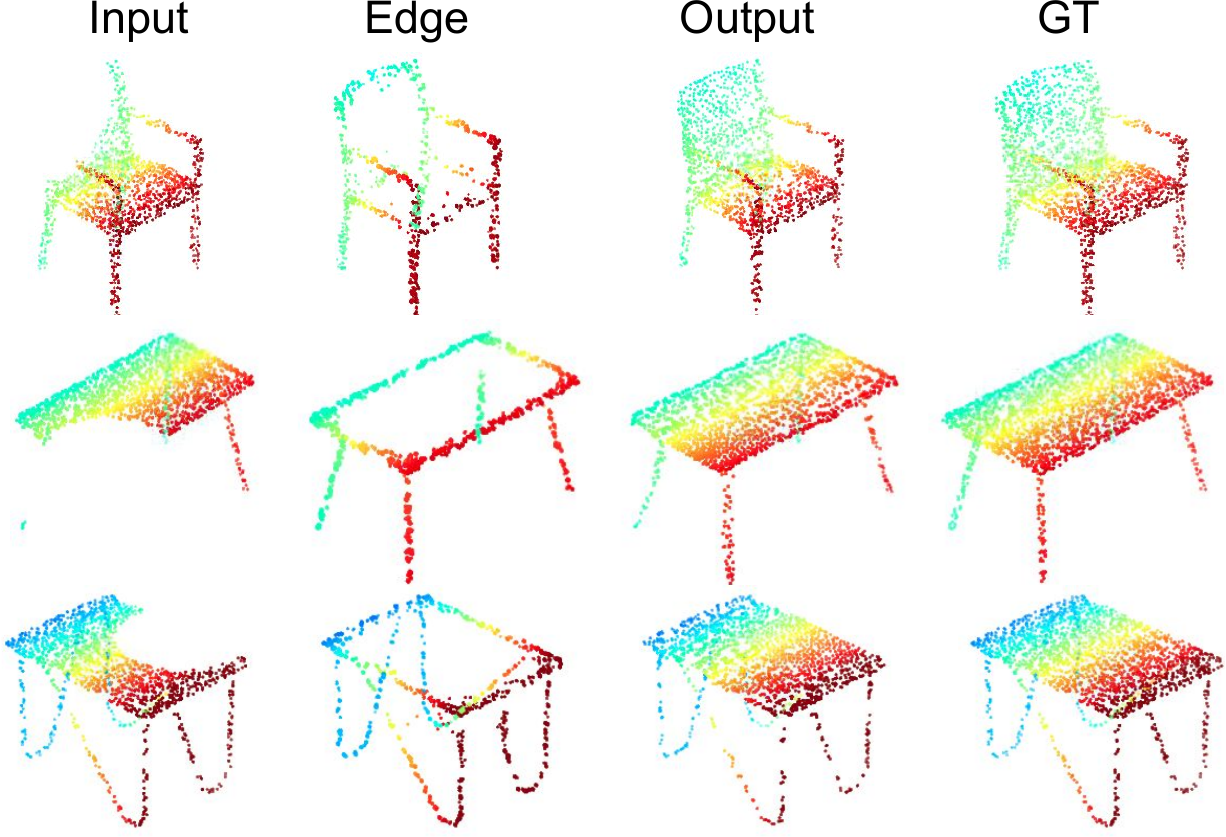}
  \vspace{-7mm}
  \caption{We propose an edge-guiding and voxel-based point cloud completion network to reconstruct complete points from incomplete inputs. Under the guidance of hallucinated edges and the help of gridding structures, we are able to generate fine-grained details for the thin structures.}
  \label{task_picture}
  \vspace{-0.6cm}
\end{figure}

Numerous methods have attempted to achieve shape completion from different representations, \emph{e.g.} meshes~\cite{groueix2018,wang20193dn}, 
implicit fields~\cite{chibane2020implicit,liao2018deep,mescheder2019occupancy,stutz2018learning} and point clouds~\cite{yuan2018pcn,topnet2019,Wang_2020_CVPR,liu2020morphing,wen2020point,sarmad2019rl,chen2020pcl2pcl,zhang2020detail,huang2020pf,wang2020point,wang2020self,wang2020softpoolnet,xie2020grnet,han2019multi,gu2020weakly}. 
Meshes represent object shapes by a set of vertices and edges. 
Despite their ability to reconstruct complex object structures, it is difficult to change shape topologies during training due to the fixed vertex connection patterns.
Implicit fields represent shapes by signed distance functions (SDF)~\cite{osher2003signed} that are more flexible to achieve arbitrary resolutions. However, learning an accurate SDF requires a large amount of sampling for a single object.
In contrast, point clouds are concise 3D shape representations and easier to add new points during training.
Nonetheless, previous point cloud shape completion works \cite{yuan2018pcn,topnet2019} are struggling at synthesizing surface details since point clouds are unordered and challenging to control. 
To alleviate this problem, recent works~\cite{liu2020morphing,Wang_2020_CVPR,xie2020grnet} propose to add a skip connection between the partial input and the decoder and have shown better reconstruction on object details. Despite the effort, the results for complex topologies are still inferior because of unrealistic points generated on the missing part.

In view of these limitations, we propose a voxel-based shape completion network that leverages the generated object edges to recover more fine-grained details. The voxel representation has also been used in GRNet \cite{xie2020grnet}, where a Gridding and Gridding Reverse layer are proposed for conversion between unordered points and 3D grids. However, their voxel representation is only used to reconstruct low-resolution shapes, and additional multi-layer perceptrons (MLPs)~\cite{qi2017pointnet} are applied for denser point set generation. 
To fully integrate the voxel representation for completion in an end-to-end manner, we generate points for every grid cell by predicting a binary classification score that indicates a cell being empty or occupied and a probability density value that estimates the number of points in the cell inspired by~\cite{lim2019convolutional}.
This operation is differentiable everywhere and thus makes it easier to generate accurate coordinates according to various shape topologies. Moreover, the end-to-end grid strategy allows us to approximate object shapes without running into memory issues since we can sample an arbitrary number of points for each grid cell. 
In this way, we are able to use a lower voxel resolution, \emph{e.g.} $32\times 32\times32$ compared to $64\times 64\times64$ in GRNet, and concurrently achieve superior results. 
In addition, we introduce a multi-scale grid transformation module to learn critical object shapes in different resolutions, while GRNet only considers the full resolutional inputs.

Although the voxel representation is shown to be superior on calculating local features~\cite{xie2020grnet,han2017high,wang2017shape}, it is challenging to generate arbitrary thin structures for various objects. We thus further introduce an edge generator to enhance our network inspired by image inpainting~\cite{nazeri2019edgeconnect}. 
Since object structures are well-represented by their edges or contours, 
it is beneficial to incorporate the edge information when generating complete 3D shapes.
Several examples of edge generation and point cloud completion are shown in Figure~\ref{task_picture}.

We evaluate our model on both the synthetic and real-world datasets. 
Qualitative and quantitative experiments are compared against existing state-of-the-art schemes.
Our key contributions are as follows:
\begin{itemize}
    \item We design a multi-scale voxel-based network to generate fine-grained details for point cloud completion.
    
    \item We incorporate object structure information into the shape completion by leveraging edge generation.

    \item We achieve state-of-the-art performances on different point cloud completion datasets.
\end{itemize}

\section{Related work}
In this section, we discuss the recent developments of 3D learning on point cloud analysis and completion.
\subsection{Point Cloud Analysis}
The pioneering work PointNet~\cite{qi2017pointnet} proposes a MLP-based network for shape classification and segmentation. 
They successfully learn the global shape by applying a max-pooling operation on the point features. 
However, they ignore the point relationships within a local area.
In view of this limitation, PointNet++~\cite{qi2017pointnet++} proposes a hierarchical point set feature learning module for feature extraction. 
Apart from the above MLP-based method, several works~\cite{li2018pointcnn,thomas2019kpconv} adopted convolution-based operations in image processing onto point clouds. 
Although they have achieved impressive performances on classification and segmentation, they are limited to complete and clean point sets. They are thus not applicable to real-world data due to noisy and incompleteness. This motivates us to design point completion networks to improve the performances of down-stream tasks.

\begin{figure*}
\centering
  \includegraphics[width=1\linewidth]{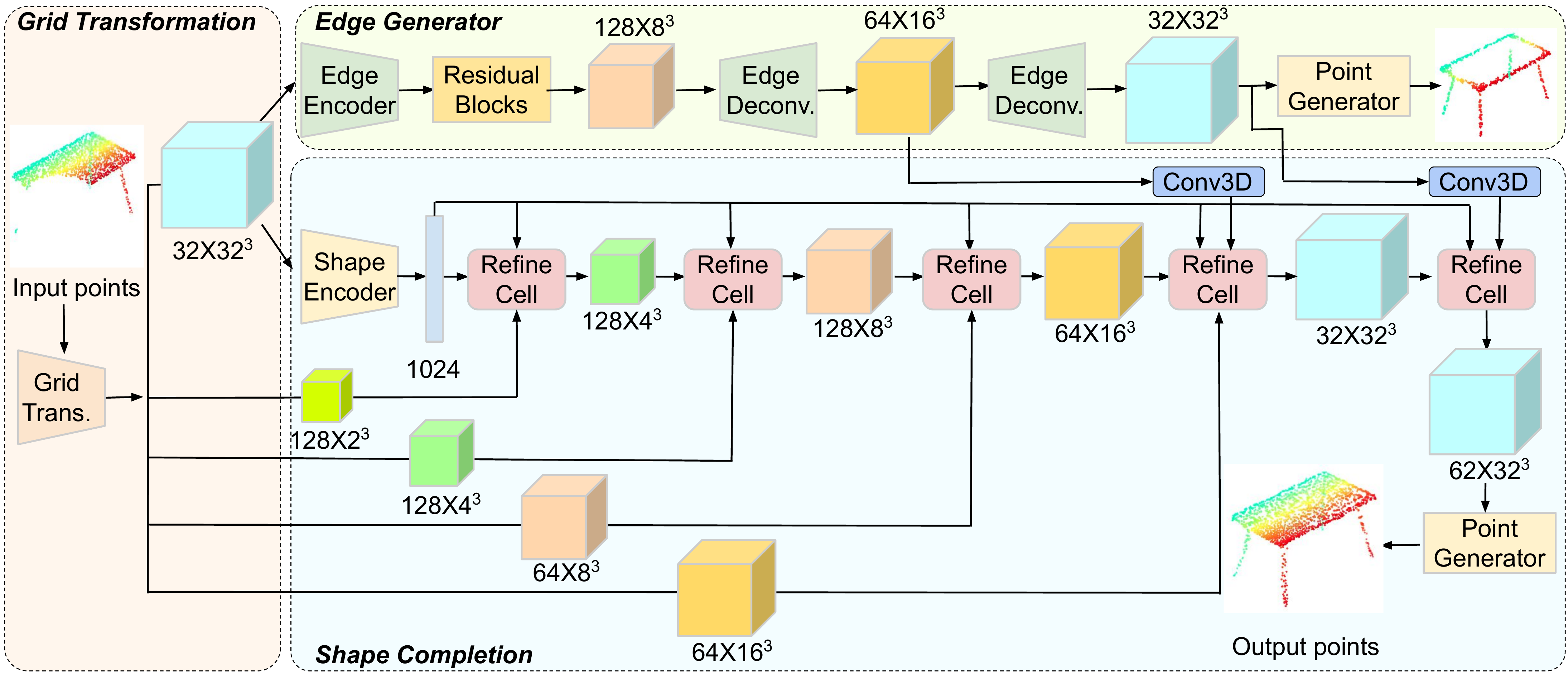}
   \vspace{-0.8cm}
  \caption{Overview of our network architecture. Given an incomplete point cloud, we first transform it to a stack of 3D features by a multi-scale grid transformation module in the left part.
  The top right branch shows the architecture of the edge generator that reconstructs edges from the grid feature $\text{P}_f^0\in R^{32\times 32^3}$. The bottom right branch presents the shape completion part, which includes a shape encoder, refinement modules and a shape decoder. The shape encoder maps the grid feature to a latent embedding $z\in R^{1024}$ that is used to project features from different scales to the output by refinement cells. The shape decoder predicts the complete point clouds by a sampling strategy.
  }
  \vspace{-6mm}
  \label{whole_network}
\end{figure*}

\subsection{Point Cloud Completion}
We roughly categorize point cloud completion into three classes: fully supervised methods, semi-supervised methods and self-supervised methods. 
The majority of works~\cite{yuan2018pcn,topnet2019,Wang_2020_CVPR,xie2020grnet,NEURIPS2020_ba036d22} follow the fully supervised manner. 
PCN~\cite{yuan2018pcn} proposes an encoder-decoder architecture for point cloud completion.
A following work TopNet~\cite{topnet2019} shares the same encoder architecture with PCN and proposes a tree-structure decoder to represent different object parts by several tree branches.
Although PCN and TopNet have achieved good performances in recovering the object shapes, they are incapable of synthesizing shape details. 
In view of this, MSN~\cite{liu2020morphing} and CRN~\cite{Wang_2020_CVPR} proposes to preserve the details of the objects from partial inputs and refine the geometric structures for the missing regions.
Following work GRNet~\cite{xie2020grnet} proposes a similar gridding network for dense point reconstruction. We differ from them as follows:
1) They use differentiable gridding and gridding reverse operations for conversion between points and grids, while our method adopts a different conversion approach and also proposes an edge generation pipeline to improve the completion performance.
We also consider partial features from different scales during shape completion (\S3.2).
2) GRNet can only obtain coarse outputs by their voxelization strategy and generate dense results by MLPs. 
In contrast, our method is able to directly generate dense and complete points in an end-to-end manner.
Another work SK-PCN~\cite{NEURIPS2020_ba036d22} proposes a similar thought that adopts skeleton generations to help the shape completion. 
However, our edges are different from their meso-skeleton. 
Their meso-skeleton focus on the overall shapes, while our edges focus on high frequency components (\emph{e.g.} thin structures), which are difficult to generate in existing methods. 
Moreover, SK-PCN generates the complete points by learning displacements from skeletal points with a local adjustment strategy, while we synthesize the complete points by injecting the edge features into the completion decoder with a multi-scale voxelization strategy. 

Instead of training the completion models using the fully complete ground truth, some works~\cite{gu2020weakly,chen2020pcl2pcl,wang2020self} propose semi-supervised or self-supervised methods to generate complete points. 
Gu $et \;al.$~\cite{gu2020weakly} propose a weakly supervised completion method, which estimates 3D canonical shapes and 6-DoF poses simultaneously from multiple partial observations. They infer the missing regions by optimizing multi-view constraints. 
To further alleviate the labeling problem, Chen $et \;al.$~\cite{chen2020pcl2pcl} propose an unpaired point cloud completion method by the adversarial training. They use generative adversarial networks (GAN)~\cite{goodfellow2014generative} to reconstruct the complete objects for the real-world partial scans. Despite their impressive performance, they need to pre-train two auto-encoders and another GAN on the latent embeddings. 
Wang $et \;al.$~\cite{wang2020self} propose a self-supervised approach for the point cloud completion, which consists of two self-training strategies when the ground truth training data are not available. 
Although we train our model with a supervised manner, we also evaluate it on the unseen categories to show our generalizability.

\section{Our Method}

\subsection{Overview}
Our objective is to reconstruct complete and high-quality 3D objects $\text{P}^\prime$ given sparse and corrupted point clouds P. Figure~\ref{whole_network} shows the illustration of our pipeline.
Most previous works~\cite{yuan2018pcn,Wang_2020_CVPR,liu2020morphing} process point sets by MLP-based neural networks with a coarse-to-fine manner and are struggling to reconstruct object details, which mainly due to two reasons: 
1) the coarse outputs created from global embeddings lose the high-frequency information for 3D objects; 2) the second stage acts as a point upsampling function that is also incapable of synthesizing complex topologies. 
To circumvent these problems, we first design an edge generator to generate the object edges. These generated edges which encode the high-frequency information of an object are then used to help the completion network to generate object details.
We further propose to adopt grid representations for both the edge generation and shape completion in view of the success of voxel-based techniques~\cite{xie2020grnet,han2017high,wang2017shape} on
local feature embedding.
We consider multi-scale voxel features to enhance the learning of shape structures. 
Overall, our point completion network consists of three stages: 1) a multi-scale grid transformer to project raw points into voxel representations, 2) an object edge generator to obtain shape edges, and 3) a shape completion network to generate complete point sets.
Detailed network architectures are shown in our supplementary materials.

\subsection{Multi-scale Grid Transformation}\label{grid_transformation}
\begin{figure}
  \includegraphics[width=\linewidth]{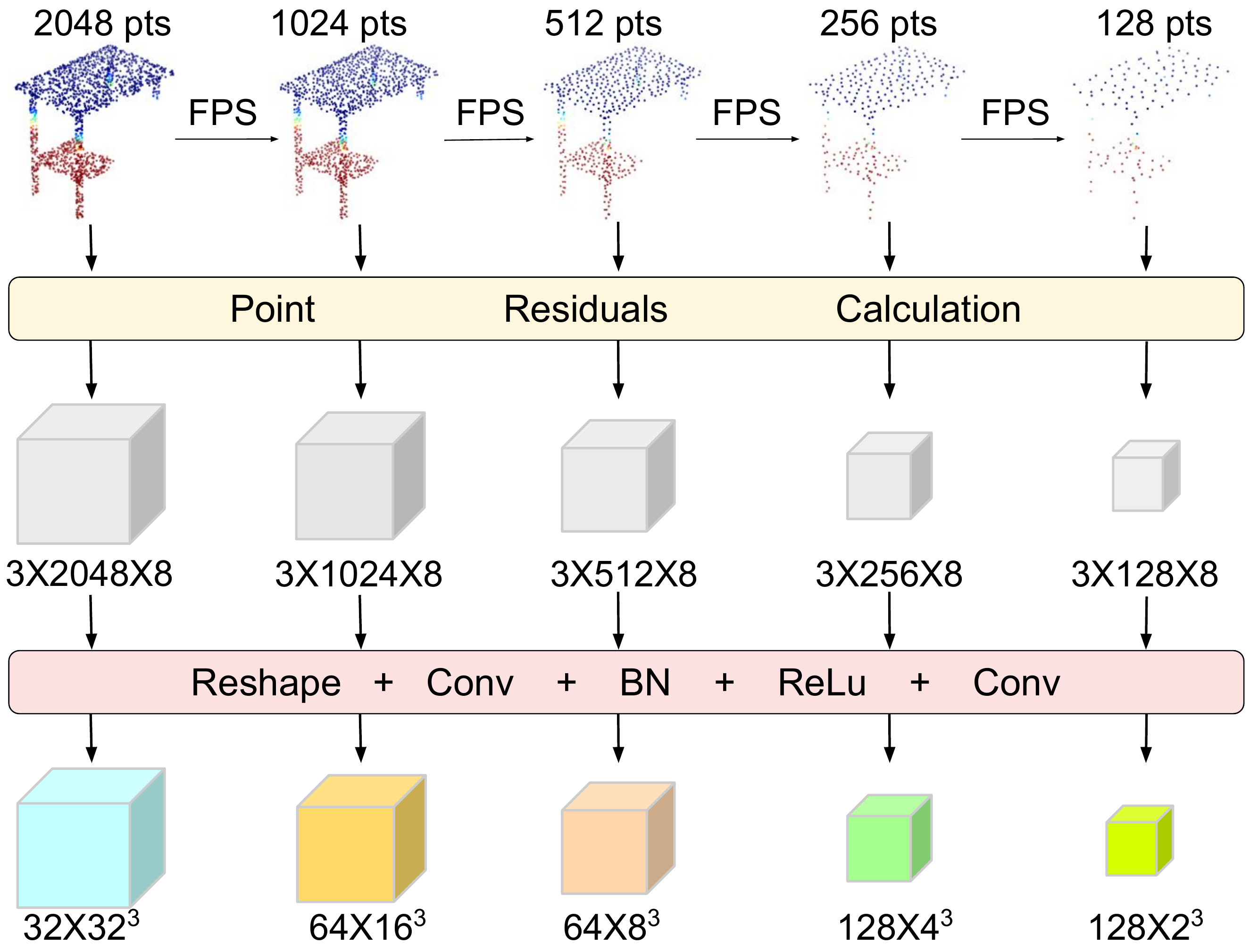}
  \vspace{-7mm}
  \caption{Illustration of the multi-scale grid transformation architecture. FPS refers to farthest point sampling~\cite{qi2017pointnet++}.}
  \vspace{-6mm}
  \label{grid_trans}
\end{figure}

Figure~\ref{grid_trans} gives the detailed illustration of the multi-scale grid transformation module shown on the left of Figure~\ref{whole_network}.
The grid transformer is used to project the partial points into 3D grids for the edge generator and shape completion network.
Instead of directly obtaining the binary representations that lead to the loss of structure information in previous works~\cite{dai2017shape}, we learn a stack of grid features $\text{P}_f^i, \{i=0, 1, 2, 3, 4\}$ by a multi-scale grid transformation module.
Specifically, we first downsample the raw partial point cloud into another four sparser points sets with a downsampling ratio of 2. 
We then calculate all the point offsets between the point coordinates and their eight nearby grid vertexes. This results in a tensor with the size of $3\times \text{N} \times 8$, where N is the point number. 
Subsequently, two 3D convolutions with batch normalization~\cite{ioffe2015batch} and ReLU activation between them are used to obtain five grid features $\text{P}_f^i$. 
The five features represent the shape features from different point resolutions.
We take mean values for all the point features that lie in the same grid as the per-grid representations. 
The size of the five grid features $\text{P}_f^i$ are $32\times32^3$, $64\times16^3$, $64\times8^3$, $128\times4^3$ and $128\times2^3$, respectively.

\subsection{Edge Generator}\label{edge_generation}
The top right of Figure~\ref{whole_network} shows our encoder-decoder structure of the edge generator.
The ground truth edges are obtained by a light-weight algorithm~\cite{ahmed2018edge}.

\subsubsection{Edge Encoder}
We extract high dimensional voxel features $\text{E}_f$ in the edge encoder by two 3D convolutional blocks that downsample $\text{P}_f^0$ by two times. 
Each convolutional block includes one 3D convolution, instance normalization (IN)~\cite{ulyanov2016instance} and ReLU.
Given the success of residual architecture~\cite{he2016deep} in deep learning, we refine the edge features $\text{E}_f$ by three residual blocks. Each block consists of two $3\times3\times3$ convolutional layers and INs. ReLU activation is added after the first IN layer. 

\subsubsection{Edge Decoder}
The edge decoder includes two deconvolutional layers followed by one IN and ReLU activation.
It upscales the feature size by a factor of 2 each time.
Another convolution is used to obtain the binary prediction score for each grid cell.
The outputs of the edge decoder used in the shape completion module are two grid features, $i.e.$, $\text{P}_e^{16}\in R^{1\times 16^3}$ and $\text{P}_e^{32}\in R^{1\times 32^3}$.

\subsubsection{Point Generator}
The point generator aims at generating point clouds from the grid representations, \emph{e.g.} $\text{P}_e^{16}$ and $\text{P}_e^{32}$.
We assume the local shape within a voxel can be approximated by a surface patch~\cite{yang2018foldingnet,groueix2018} and thus adopt the folding mechanism to repeat the point features within a grid cell to obtain numerous point features $f_g$.
Consequently, we generate multiple points for every voxel by predicting $p_c$ and $\delta_c$ from $f_g$ for each grid cell instead of generating one point for each cell~\cite{xie2020grnet} inspired by~\cite{lim2019convolutional}.
$p_c$ is a non-zero classification score indicating whether the cell is occupied or empty. 
$\delta_c$ is the density value indicating the number of points in a given cell.
We predict the residual offsets for each grid cell, and the final complete point clouds $\text{P}^\prime$ are obtained by adding the offsets on the corresponding cell centers.

\subsection{Shape Completion}
The shape completion network is shown in the bottom right of Figure~\ref{whole_network}, which includes a shape encoder, an iterative refinement module and a point generator.
The inputs of the shape completion are the partial grid features $\text{P}_f^i$ and the edge representations $\text{P}_e^{16}$ and $\text{P}_e^{32}$.
The completion synthesis begins at a small resolution of $2^3$, and grid features are iteratively refined with the help of Adaptive Instance Normalization (AdaIN)~\cite{huang2017arbitrary,lim2019convolutional}. 

\subsubsection{Shape Encoder}
The shape encoder is used to extract global embeddings $z\in R^{1024}$ from the partial features.
Given the highest resolutional spatial feature $\text{P}_f^0\in R^{32\times32^3}$, four partial grid features $\text{C}_f^i$ are obtained by four 3D convolutional blocks. The number of feature channels are doubled and the spatial size are downsampled by a factor of $2^3$ as $i$ increases from 0 to 3 in different blocks.
These four convolution blocks consist of two 3D convolutions and one batch normalization and a max-pooling layer. 
$z\in R^{1024}$ is calculated by downsampling the feature $\text{C}_f^3\in R^{2^3\times512}$ by another convolutional block followed by a reshape operation.
This block downsamples the spatial size by setting the padding value to 0 in the 3D convolution. 

\subsubsection{Refinement Modules}
Refinement modules aim to refine and upsample the grid features during the shape completion process.
Every refinement cell operates at a specific spatial size and comprises a series of upsampling, 3D convolution, instance normalization, affine feature transformation and ReLU activation.  
The input $x$ for each refinement cell includes three elements: the output from the previous layer, the partial point feature $\text{P}_f^i$ obtained from the grid transformation module and the binary edge features calculated from the edge generator. They are concatenated on channel dimensions.
Unlike GRNet~\cite{xie2020grnet} which only considers the global information in the first layer of the decoder, we incorporate the global embedding in all the subsequent layers of the decoder by AdaIN. We first calculate the channel-wise mean $\mu(y)$ and variance $\sigma(y)$ for input $x$ from the global embedding $z$ by one fully connected layer.
The input features $x$ are then projected by $\text{AdaIN}(x,y)=\sigma(y)(\frac{x-\mu(x)}{\sigma(x)})+\mu(y)$.
$\mu(x)$ and $\sigma(x)$ are the channel-wise mean values and variances of input $x$.
The overall number of cascaded refinement modules depends on the output resolution, which is five for the resolution of $32\times32\times32$ in our work.
We generate object points from the grid features by the same point generator in edge generation with different parameters. 

\subsection{Optimization}
Since our edge generator and point completion network are both differentiable, our entire voxel-based network can be optimized in an end-to-end manner. 
The overall constraints consist of the losses for the complete outputs and edge results. 

We calculate the Chamfer Distance (CD)~\cite{fan2017point} and its sharper version~\cite{lim2019convolutional} on the output point clouds $\text{P}^\prime$ and the ground truth points Q as:
\begin{align}
    \mathcal{L}_{\text{CD}}&=\frac{1}{\text{N}}\sum_{p\in \text{P}^\prime} d(p,\text{Q})^2+\frac{1}{\text{M}}\sum_{q\in \text{Q}}d(q,\text{P}^\prime)^2\\
    \mathcal{L}_{\text{CD}}^{\text{S}}&=\frac{1}{\text{N}}(\sum_{p\in \text{P}^\prime} d(p,\text{Q})^5)^{\frac{1}{5}}+\frac{1}{\text{M}}(\sum_{q\in \text{Q}}d(q,\text{P}^\prime)^5)^{\frac{1}{5}},
\end{align}
where N and M are the numbers of the outputs and ground truth points, respectively. Additionally, the Chamfer Distances $\mathcal{L}_{\text{CD}}^{E}$ between the generated edge points and the ground truth edges are computed.

We also compute the binary cross entropy (BCE) loss between the predicted completion voxels and the ground truth grids as:
\begin{align}
    \mathcal{L}_{\text{BCE}}^{P}=-\frac{1}{32^3}\sum_{c} \widehat{p_c}\cdot log(p_c) + (1-\widehat{p_c})\cdot log(1-p_c),
\end{align}
where $\widehat{p_c}$ and $p_c$ are the ground truth grid probability and predicted grid probability, respectively. Similarly, $\mathcal{L}_{\text{BCE}}^{E}$ represents for the edge loss.

Since our final outputs are the offsets to the per-grid centers, we compute the density estimation and the quality of locality~\cite{lim2019convolutional} to enhance the constraint on the grid cells.
The density loss and the locality measurement are given by the mean squared error:
\begin{align}
    \mathcal{L}_d(\delta,\hat{\delta})=\frac{1}{32^3}\sum_{c}(\delta_c-\widehat{\delta_c})^2
\end{align}
and 
\begin{align}
    \mathcal{L}_o=\sum_c \sum_{p\in \text{P}^{\prime}} \text{max} (\text{dist}(p,c_o)-\sqrt{3},0).
\end{align}
$c_o$ is the cell center, $\delta_c$ and $\widehat{\delta_c}$ are the predicted grid density and ground truth density, respectively. 
We also compute $\mathcal{L}^{E}_d(\delta,\hat{\delta})$ and $\mathcal{L}_o^E$ for the edge reconstruction. 

Finally, the overall losses of our network are given by:
\begin{equation}
    \begin{aligned}
        \mathcal{L}_{all}=\lambda_1(\mathcal{L}_{\text{CD}}+\mathcal{L}_{\text{CD}}^{E})+
        \lambda_2\mathcal{L}_{\text{CD}}^{\text{S}}
        + \lambda_3 (\mathcal{L}_{\text{BCE}}^{P}+\mathcal{L}_{\text{BCE}}^{E})\\
        +\lambda_4(\mathcal{L}_d(\delta,\hat{\delta})+\mathcal{L}^{E}_d(\delta,\hat{\delta}))+
        \lambda_5(\mathcal{L}_o+\mathcal{L}_o^E),
    \end{aligned}
\end{equation}
where $\lambda_i$ are weights to balance the different loss terms. 

\section{Experiments}
\subsection{Datasets}

We first evaluate our method on the widely used benchmarks of 3D point cloud completion, \emph{e.g.} Completion3D~\cite{topnet2019} and the PCN dataset~\cite{yuan2018pcn}, which are large-scale datasets derived from the ShapeNet dataset. We follow the settings of training, validation and testing splits in the Compeltion3D and PCN dataset for fair comparisons. 
The incomplete points are obtained by back-projecting 2.5D depth images from a partial view into the 3D space, and the complete points are uniformly sampled from the mesh models. These two datasets contains eight categories: airplane, cabinet, car, chair, lamp, sofa, table and vessel.
The results on the Completion3D benchmark are directly obtained from the benchmark~\footnote{https://completion3d.stanford.edu/results} and the results of the PCN dataset are shown in the supplementary materials. 

\input{tables/quantative_topnet}

To further test the generalization ability on more categories and the capability of reconstructing more complex structures, we create a new dataset consisting of 12 categories from the ShapeNet dataset instead of 8 classes in the Completion3D and PCN datasets. 
The objects include bag, cap, car, chair, earphone, guitar, lamp, laptop, motorbike, mug, skateboard and table.
There are 12,137 training data, 1,870 validation data and 2,874 testing data, respectively.
The ground-truth point clouds are obtained by uniformly sampling 2,048 points from 3D meshes.
Instead of creating the partial views following the PCN dataset, we explore the other widely used method~\cite{huang2020pf,sarmad2019rl,richard2020kaplan} to randomly select a viewpoint as a center and remove points within a certain radius from complete data to obtain the partial inputs. This is to show the generalization ability and robustness of our algorithm on another type of incompleteness.

We also test on unseen categories of our dataset and the real-world KITTI dataset~\cite{geiger2013vision} to verify the robustness of our method. 

\input{tables/quantative_seen}

\begin{figure}
    \centering
  \includegraphics[width=\linewidth]{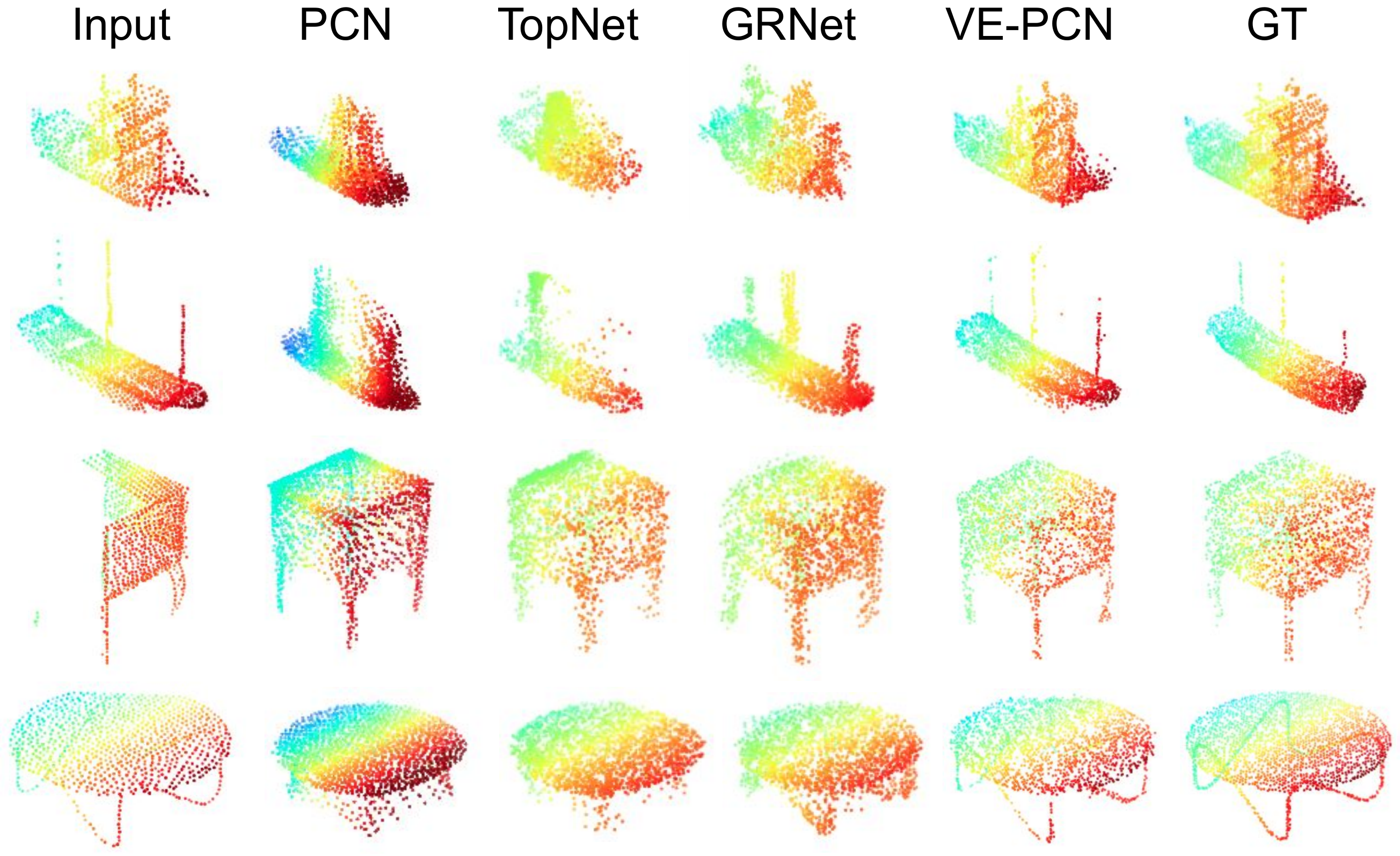} 
  \vspace{-5mm}
  \caption{Qualitative results on the Completion3D dataset.}
  \label{topnet_qulitative}
  \vspace{-5mm}
\end{figure}

\begin{figure*}
    \centering
  \includegraphics[width=\linewidth]{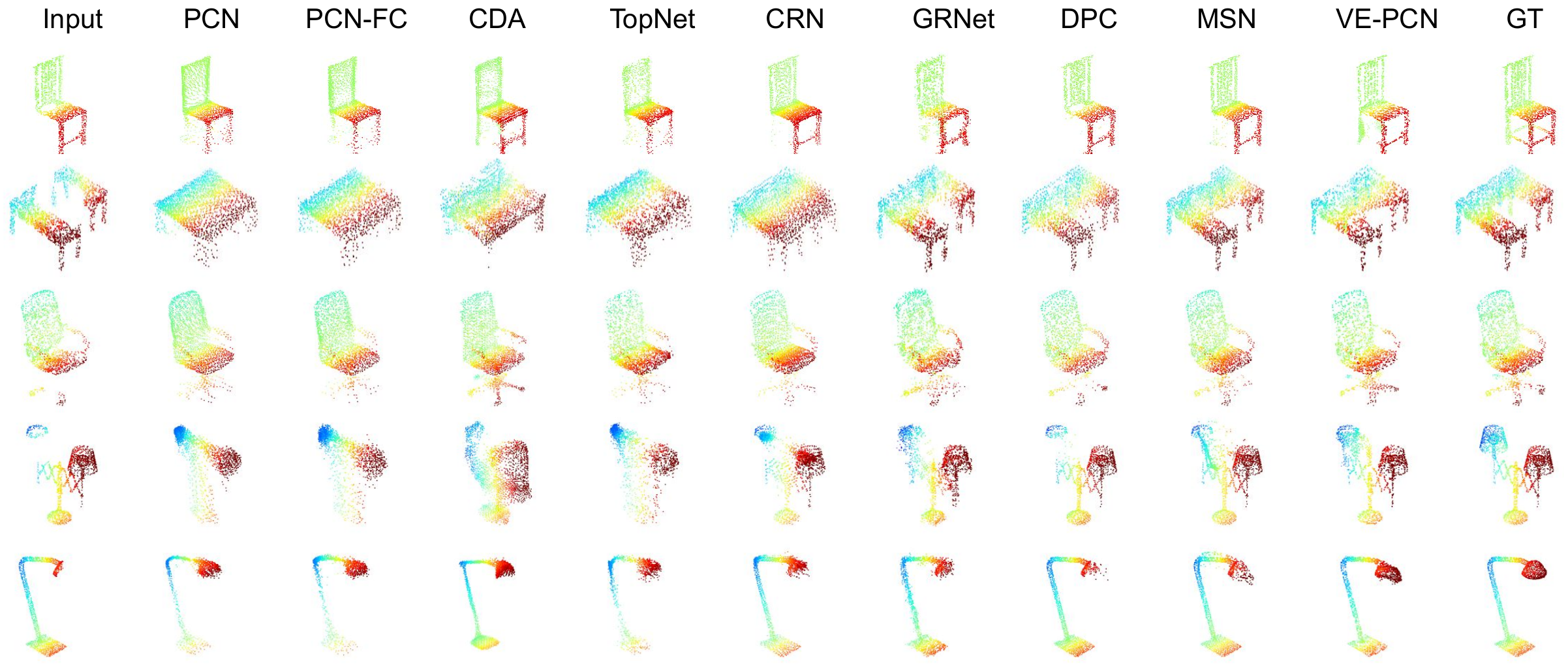} 
  \vspace{-7mm}
  \caption{Qualitative comparisons of point cloud completion on seen categories of our dataset.}
  \label{qualitative_seen}
  \vspace{-4mm}
\end{figure*}

\subsection{Implementation Details}
We empirically set the weights of the loss terms to $\lambda_1=1e4$, $\lambda_2=300$, $\lambda_3=100$,  $\lambda_4=1e10$, $\lambda_5=0.3$ for the Completion3D dataset and our dataset. $\lambda_2=\lambda_5=0$ for the PCN dataset.
We train our model with a learning rate of 0.0007 and a batch size of 32 on one TITAN X GPU.

\subsection{Completion3D Dataset}
We compare the quantitative results with existing point cloud completion methods in Table~\ref{quantative_topnet}, where our method VE-PCN achieves the best performance in terms of the average Chamfer Distance across all the categories. 
Compared with existing state-of-the-art approaches, we achieve better results in six categories, which verifies the effectiveness of our method. 
The superior performances of object details in Figure~\ref{topnet_qulitative} prove the superiority of our edge-guiding strategy and voxelization technique. 

\subsection{Our Dataset}
\noindent\textbf{Seen Categories.}
Qualitative and quantitative results are shown in Figure~\ref{qualitative_seen} and  Table~\ref{quantative_seen}, respectively.
We retrain other models on our dataset using their released codes.
The number of output points is 2,048, and all the object categories are seen during training. 
Apart from CD, we adopt Fr$\acute{\text{e}}$chet Point Cloud Distance (FPD)~\cite{Shu_2019_ICCV} as another evaluation metric.
As shown in Figure~\ref{qualitative_seen}, previous approaches show inferior performances on reconstructing the missing parts or fine-grained object details, while our model synthesizes more realistic object shapes with accurate surfaces.  
For example, other methods fail to reconstruct the object details of the lampshade (the last second row). In contrast, our method is able to preserve lamp details in the partial input and concurrently generate details for the missing regions. 
Other examples such as the thin legs of the chair and table further verify this.
The quantitative results shown in Table~\ref{quantative_seen} 
verify that we achieve superior scores in the coordinates approximation and distribution estimation.
We decrease CD error 
by around  $24.8\%$
compared to the second-best result and achieve the best performances on almost all the categories compared to other approaches. 
Overall, the superior qualitative and quantitative performances verify the effectiveness of our edge generation and voxelization strategies for point cloud completion. 
As shown in Table~\ref{quantative_diff_resolutions}, we also obtain better results with different resolutions. 
This indicates that our edge-guiding network and grid representations are more robust to different kinds of completion scenarios.
More results are shown in supplementary materials.
\input{tables/quantative_diff_resolutions}

\noindent\textbf{Unseen Categories.}
Table~\ref{quantative_unseen} shows the results on unseen categories, which contain airplane, knife, pistol and rocket.
We directly use the models trained on the seen categories for testing.
Our method achieves lower CD errors compared to other state-of-the-art works and synthesizes high-fidelity object shapes even though our network is not trained on those categories.
We obtain $18.4\%$ relative improvement compared to the second-best method GRNet~\cite{xie2020grnet}, which demonstrates that our method has better generalization.
\input{tables/quantative_unseen}
\input{tables/kitti_quantitative}

\vspace{-1mm}
\subsection{KITTI Dataset}
\begin{figure}
    \centering
    \vspace{-2mm}
  \includegraphics[width=0.95\linewidth]{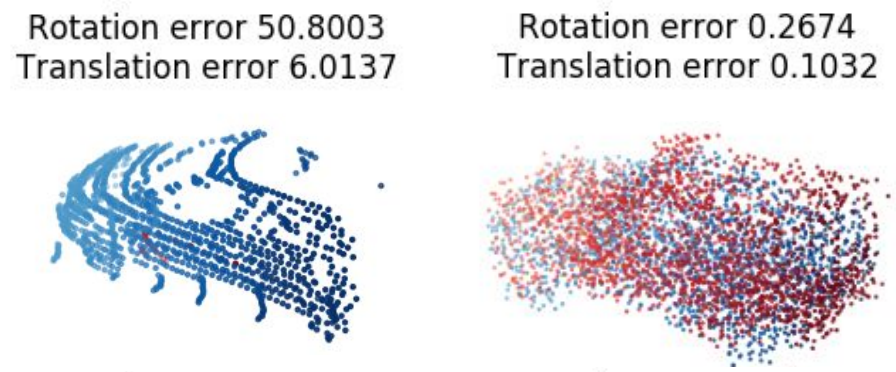}
  \vspace{-5mm}
  \caption{Qualitative comparisons on car generation from the KITTI dataset.}
  \label{kitti_qualitative}
  \vspace{-6mm}
\end{figure}

We further compare the robustness of our method with other approaches on the real-world KITTI~\cite{geiger2013vision} dataset. 
We directly adopt the models trained on the car category of our dataset for testing since there are no ground truths for KITTI. 
We make use of the fidelity distance as the evaluation metric following~\cite{yuan2018pcn,xie2020grnet,zhang2020detail}, which measures how well the inputs are preserved in the outputs.
Table~\ref{kitti_quantitative} indicates that our method is capable of reconstructing more realistic cars, although the examples are noisy and severely occluded. 
The result of our model without edge generation is 0.0279, which is worse than our entire pipeline. This verifies that our edge guidance is beneficial for real-world datasets.
We also measure the registration errors between neighboring frames in the same Velodyne sequence following ~\cite{yuan2018pcn}. 
An example in Figure~\ref{kitti_qualitative} shows that our method can decrease the rotation and translation errors by completing the points when compared to the rotation and translation errors from raw inputs.
More results are in supplementary materials.

\subsection{Edge Generation}
\begin{figure}
    \centering
  \includegraphics[width=\linewidth]{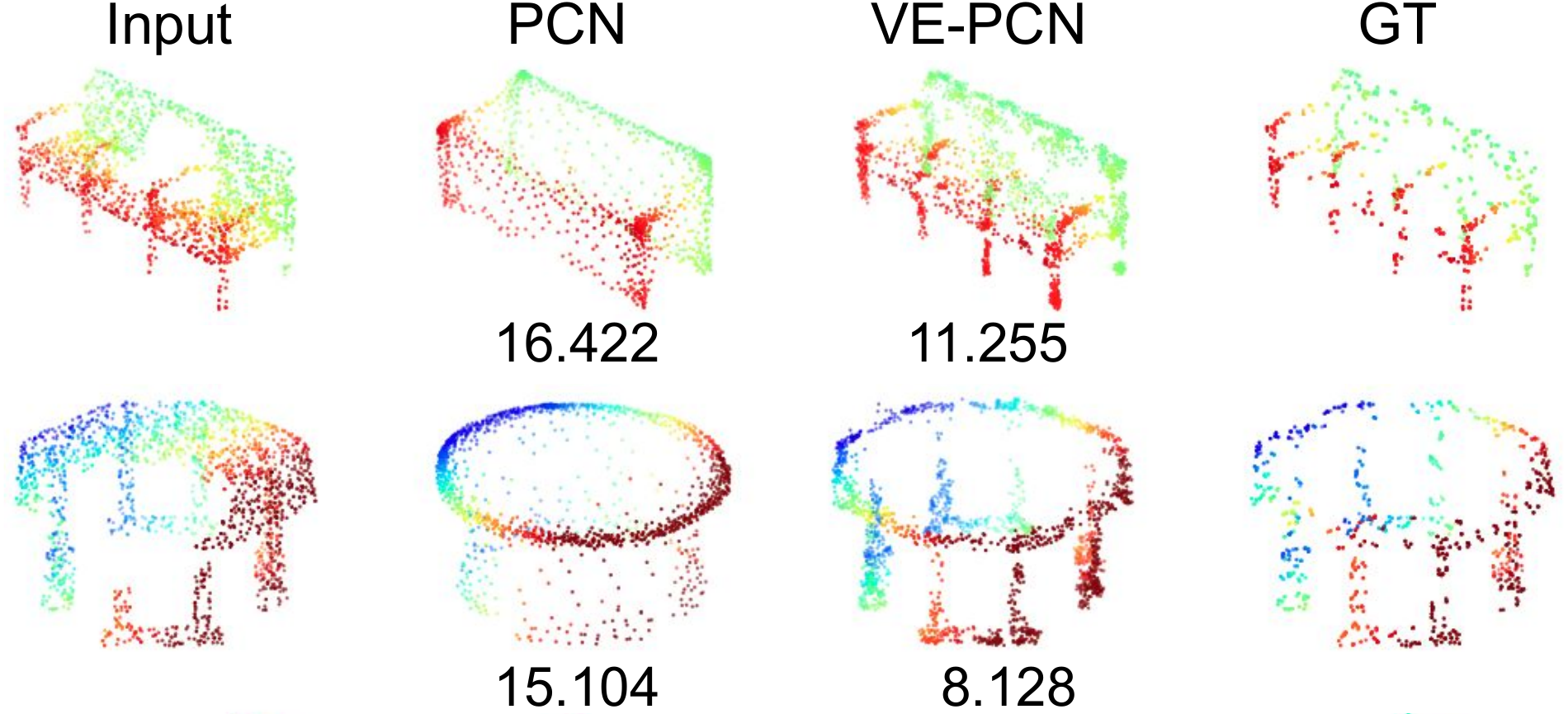}
  \vspace{-7mm}
  \caption{Comparisons of object edge generation with PCN. Bottom values are the CD errors per point ($10^{-4}$).}
  \label{edge_qualitative}
  \vspace{-7mm}
\end{figure}

We compare our voxel-based method to the classical MLP-based method PCN~\cite{yuan2018pcn} for the point cloud edge generation and show the results in Figure~\ref{edge_qualitative}.
Our edges are cleaner and more accurate compared to the results of PCN,
and our output edges cover all the thin structures of an object, while the results from PCN are over-smoothed.
This demonstrates that grid representation is desirable for the reconstruction of thin structures.
More results are shown in supplementary materials.

\subsection{Ablation Studies}
This section explores the effects of different network modules by removing or replacing a specific component. The quantitative results are reported in Table~\ref{ablation}.
The training of ablation experiments is conducted on our created ShapeNet dataset with an output resolution of 2,048. 

MGT represents our multi-grid transformation module. DG represents the differentiable Gridding module in~\cite{xie2020grnet}. We replace our MGT with DG for comparison. 
We further test our model without MGT, where we directly feed the binary voxelizations obtained from raw points to the edge generator and shape completion network.
EG is the edge generator and PE represents partial edges directly calculated from~\cite{ahmed2018edge}. We replace EG with PE when using PE.

As shown in Table~\ref{ablation}, the performance drops upon the removal of our MGT (ID: 3 vs 5) or the use of DG~\cite{xie2020grnet} (ID: 3 vs 4). This indicates that our multi-scale grid feature learning plays an essential role in edge generation and shape completion.
Additionally, we obtain worse results by either removing the edge generator (ID: 1 vs 3) or replacing it with the calculated partial edges (ID: 1 vs 2). This verifies the importance of complete edges hallucinated from our edge generator.
More ablation studies on different losses are shown in the supplementary materials.
\input{tables/ablation}

\section{Conclusion}
We present a voxel-based network for point cloud completion by leveraging the edge generation. To reconstruct the realistic structures with fine-grained details, we propose to generate the object edges and complete point sets with the guidance of the hallucinated edges. Furthermore, we transform the unordered point sets into grid representations to support edge generation and point cloud reconstruction.
Our multi-scale grid feature learning further boost the shape completion performance, and we obtain superior performance on different point cloud completion datasets. 

\vspace{-4mm}
\paragraph{Acknowledgments.}
This work is supported in part by the Singapore MOE Tier 2 grant MOE-T2EP20120-0011, and the National Research Foundation, Prime Ministers Office, Singapore, under its CREATE programme, Singapore-MIT Alliance for Research and Technology (SMART) Future Urban Mobility (FM) IRG.

\clearpage


\twocolumn[{%
		\renewcommand\twocolumn[1][]{#1}%
		\vskip .5in
		\begin{center}
			\textbf{\Large Voxel-based Network for Shape Completion by Leveraging Edge Generation -}\\
			\vspace*{6pt}
			\textbf{\Large  Supplementary Material}\\
			\vspace*{10pt}
			{\large
				Xiaogang Wang\quad\quad Marcelo H Ang Jr\quad\quad Gim Hee Lee\\
			}
			\vskip .5em
			{\large National University of Singapore\\}
			{\tt\small xiaogangw@u.nus.edu\quad \{mpeangh,gimhee.lee\}@nus.edu.sg}
			\vspace*{10pt}
		\end{center}
	}]

\setcounter{equation}{0}
\setcounter{figure}{0}
\setcounter{table}{0}
\setcounter{section}{0}

\section{Ground Truth Edge Generation}
We obtain the ground truth edges $\widehat{\text{P}_e}$ with~\cite{ahmed2018edge}.
Specifically, object edges are identified by evaluating the query point $p$ from its k-nearest neighbors. 
We first find the k-nearest neighbors 
of each query point from the object and denote $c$ as the center of these neighbors. 
We then calculate the minimum distance $v$ among all the neighboring points to the query point. A query point $p$ is classified as the edge point if $||c-p||>\lambda \cdot v$. 
We set $\lambda=5$ and $k=100$ for our created dataset, and set $\lambda=1.8$ and $k=150$ for the Completion3D dataset.

\input{tables_supp/model_size}
\input{tables_supp/ablation_supp}
\input{tables/pcn_results}
\input{tables/quantative_seen_fpd}
\input{tables_supp/quantitative_occlusion}

\section{Evaluation Metrics}
Following previous methods~\cite{yuan2018pcn,topnet2019,Wang_2020_CVPR,xie2020grnet},
we use the CD and Fr$\acute{\text{e}}$chet Point Cloud Distance (FPD)~\cite{Wang_2020_CVPR,Shu_2019_ICCV} as the evaluation metrics for the synthetic datasets and fidelity and registration errors for the KITTI dataset.

\paragraph{FPD.}
FPD evaluates the distribution similarity by
the 2-Wasserstein distance between the real and fake Gaussian 
measured in the feature spaces of the point sets, \ie
\begin{equation}
    \begin{split}
    \text{FPD}(\text{X},\text{Y})=\|\text{m}_{\text{X}}-\text{m}_\text{Y}\|_2^2 +
    \text{Tr}(\Sigma_{\text{X}}+\Sigma_{\text{Y}} -2(\Sigma_{\text{X}}\Sigma_\text{Y})^{\frac{1}{2}}).
    \end{split}
\end{equation}

\paragraph{Fidelity.}
Fidelity measures the average distance from each point in the input to its nearest neighbor in the output. It evaluates how well the input points are preserved in the output. 

\paragraph{Registration Errors.}
Rotation and translation errors are the evaluation metrics for the point cloud registration. More specially, it measures the registration performances between neighboring frames in the same Velodyne sequence. 
Two types of inputs are evaluated: the partial points from the raw scans, and the generated complete points by different models.
The rotation error is computed as $2cos^{-1}(2<q_1,q_2>^2-1)$, where $q_1$ and $q_2$ are the ground truth rotation and the rotation computed from ICP, respectively.
The translation error is computed as $||t_1-t_2||_2$, in which $t_1$ is the ground truth translation and $t_2$ is the translation generated by ICP, respectively.

\section{More Details of Conversion between Points and Grids}

\paragraph{1) Conversion from points to grids.} 
We calculate the initial grid features as the coordinate differences between points and their corresponding eight nearby grid vertexes in five different scales. This results in five tensors of sizes $3\times 2048\times 8$, $3\times 1024\times 8$, $3\times 512\times 8$, $3\times 256\times 8$ and $3\times 128\times 8$, respectively.
Corresponding to the different point resolutions $\{2048, 1024, 512, 256, 128\}$, the five voxel resolutions are $\{32^3, 16^3, 8^3, 4^3, 2^3\}$.
A set of grid features $\text{P}_f^i, \{i=0, 1, 2, 3, 4\}$ are obtained from the initial grid features by several convolutional blocks (\S3.2). 
The quantitative comparison between our proposed grid transformation and the Gridding of GRNet is shown in rows 3 and 4 of Table 6 in the main paper, i.e., our transformation achieves $24.5\%$ relative improvement compared to Gridding (3.600 vs 4.768). 

\paragraph{2) Conversion from grids to points.}
GRNet predicts 262,144 ($64^3$) points from every vertex feature, and then samples 2048 points as the coarse output $\text{P}_C$ and use MLPs to generate dense points from $\text{P}_C$. 
In contrast, we directly predict the dense points by adding the point offsets to the grid centers. The number of points for each grid cell are decided by the binary score $p_c$ and density value $\delta_c$ (\S3.3.3). 

\section{Network Architecture Details}
We express the 3D convolutions with its number of output channels, the kernel size, the stride and padding values. 
For example, C3D(O1K3S1P1) indicates a 3D convolutional layer with the number of output channel as 1, kernel size as $3\times3\times3$, the stride and padding values as 1. 
DC3D represents a 3D deconvolutional layer.
We set the dilation value for all our convolutions as 1 except the first 3D convolution in the residual blocks of the edge generator.

\begin{figure*}
    \centering
  \includegraphics[width=\linewidth]{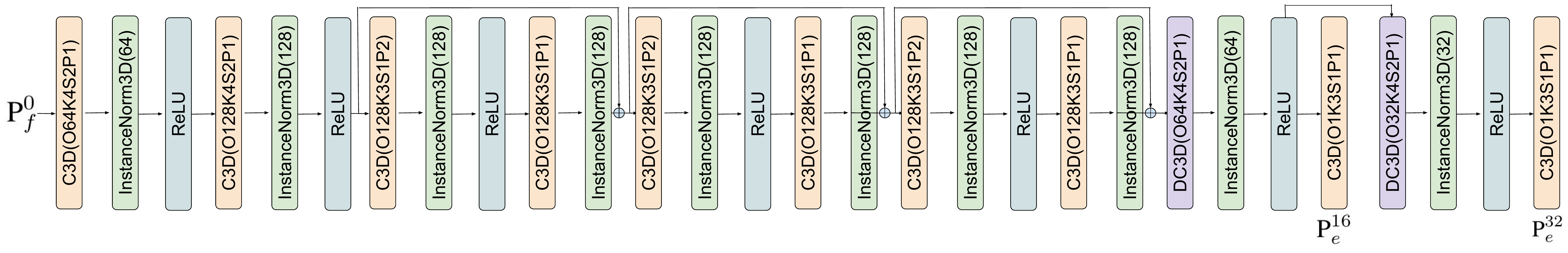}
  \caption{Architecture details of the edge generator.}
  \label{Edge_Generator}
\end{figure*}

\begin{figure*}
    \centering
  \includegraphics[width=\linewidth]{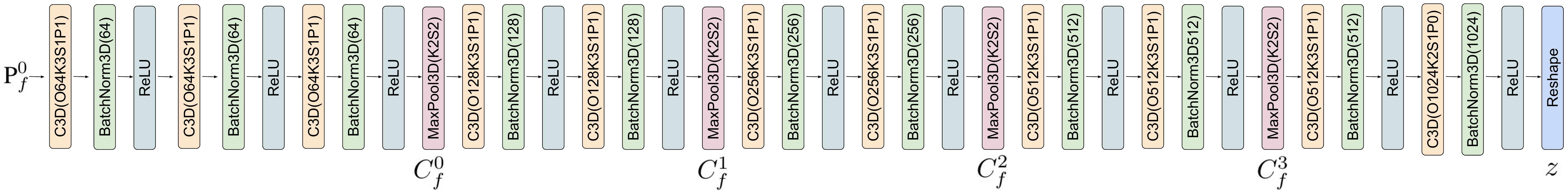}
  \caption{Architecture details of the shape encoder.}
  \label{shape_encoder}
\end{figure*}

\begin{figure*}
    \centering
  \includegraphics[width=0.6\linewidth]{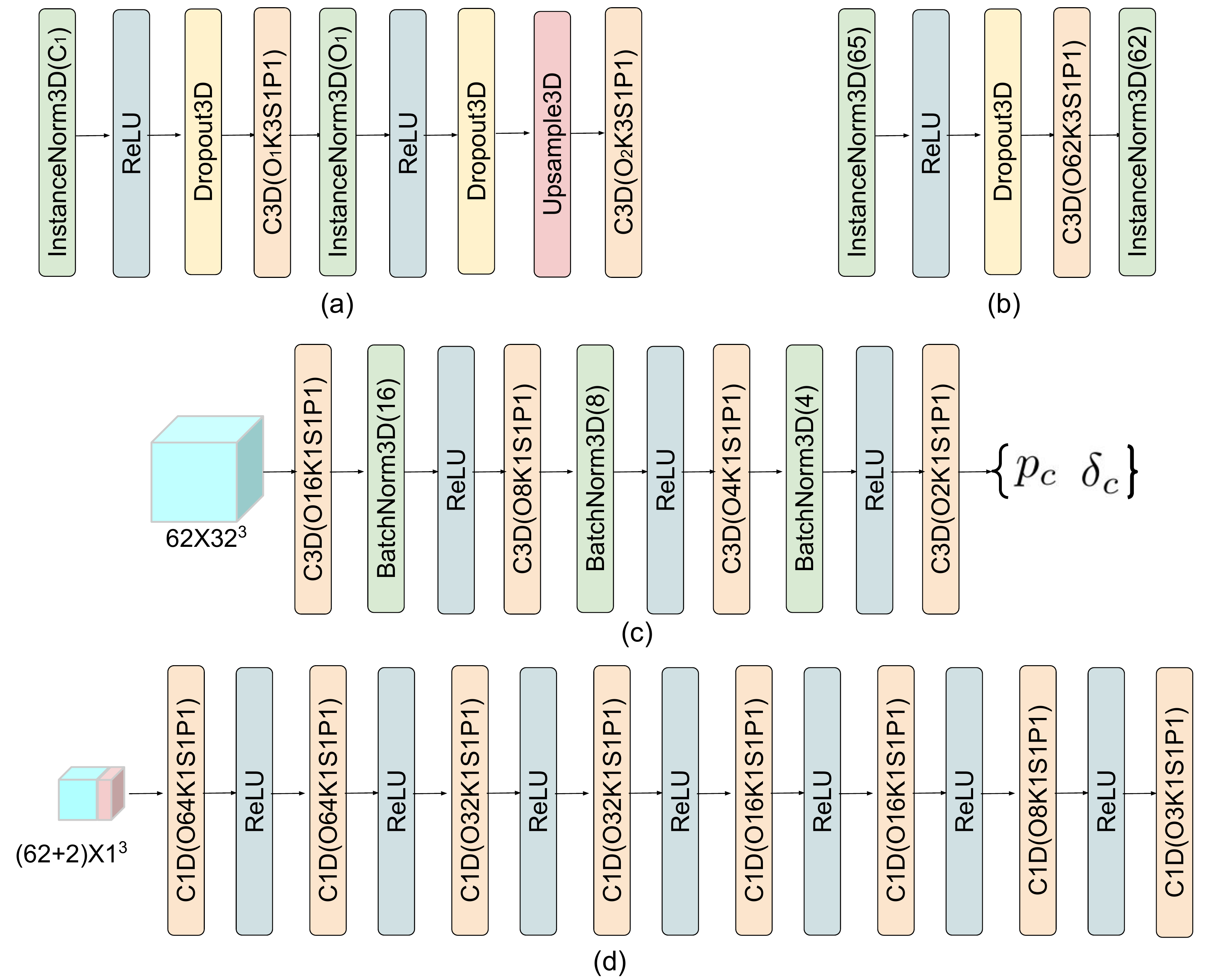}
  \caption{Architecture details of the refinement cells and shape decoder.}
  \label{refinment_decoder}
\end{figure*}

The network architectures of the edge generator and shape encoder are shown in Figures~\ref{Edge_Generator} and \ref{shape_encoder}, respectively.
Figure~\ref{refinment_decoder} shows architectures of the refinement cells and the shape decoder. 
Figure~\ref{refinment_decoder} (a) and (b) show the first four refinement cells and the last refinement cell, respectively. 
Every cell shares similar architectures but with different feature dimensions.  
$\text{C}_1$, $\text{O}_1$ and $\text{O}_2$ in the first four refinement cells are $\{128, 128, 128\}$, $\{256, 128, 128\}$, $\{192, 64, 64\}$ and $\{129, 32, 32\}$, respectively.

Figure~\ref{refinment_decoder} (c) and (d) show the architecture of point generator in the shape completion module and the edge generator.
Figure~\ref{refinment_decoder} (c) illustrates the prediction architectures for the classification score $p_c$ and density value $\delta_c$ of each grid cell. 
Figure~\ref{refinment_decoder} (d) presents the point set generation for each grid. 
Prior to feeding the grid features into the convolutional layers,
grid features are concatenated with 2 dimensional randomly sampled values to increase the point diversity in a local patch.  

\section{Time and Space Complexity Analysis}
The number of parameters and inference time of different methods are shown in Table ~\ref{model_size}. 
Some layers in our network are 1D CNNs instead of 3D CNNs (e.g. the architecture of point generator) and thus we consume much smaller parameters than voxel-based methods (CDA and GRNet). 
We compute the average inference time of 5000 forward steps on a Titan X GPU, and our time complexity is comparable to other methods.

\section{More Ablation Studies}
More ablation studies on different losses are shown in Table~\ref{ablation_supp}. 
We test the effects of $\mathcal{L}_{\text{CD}}^\text{S}$, $\mathcal{L}_{\text{CD}}$, $\mathcal{L}_o$ and $\mathcal{L}_{\text{BCE}}^\text{E}$ by setting the corresponding weights to be 0.
The results are obtained by testing on our created dataset.

\section{More Experimental Results}
\subsection{Results on the PCN Dataset}
We show the results of our method on the PCN dataset in Table~\ref{pcn_results}. All the other results are cited from the state-of-the-art work PMP-Net~\cite{wen2021pmp}.
We achieve lower average CD errors compared to all prior works and obtain better performances on the majority of object categories.

\subsection{More Results on the Completion3D Dataset}
\begin{figure*}
    \centering
  \includegraphics[width=\linewidth]{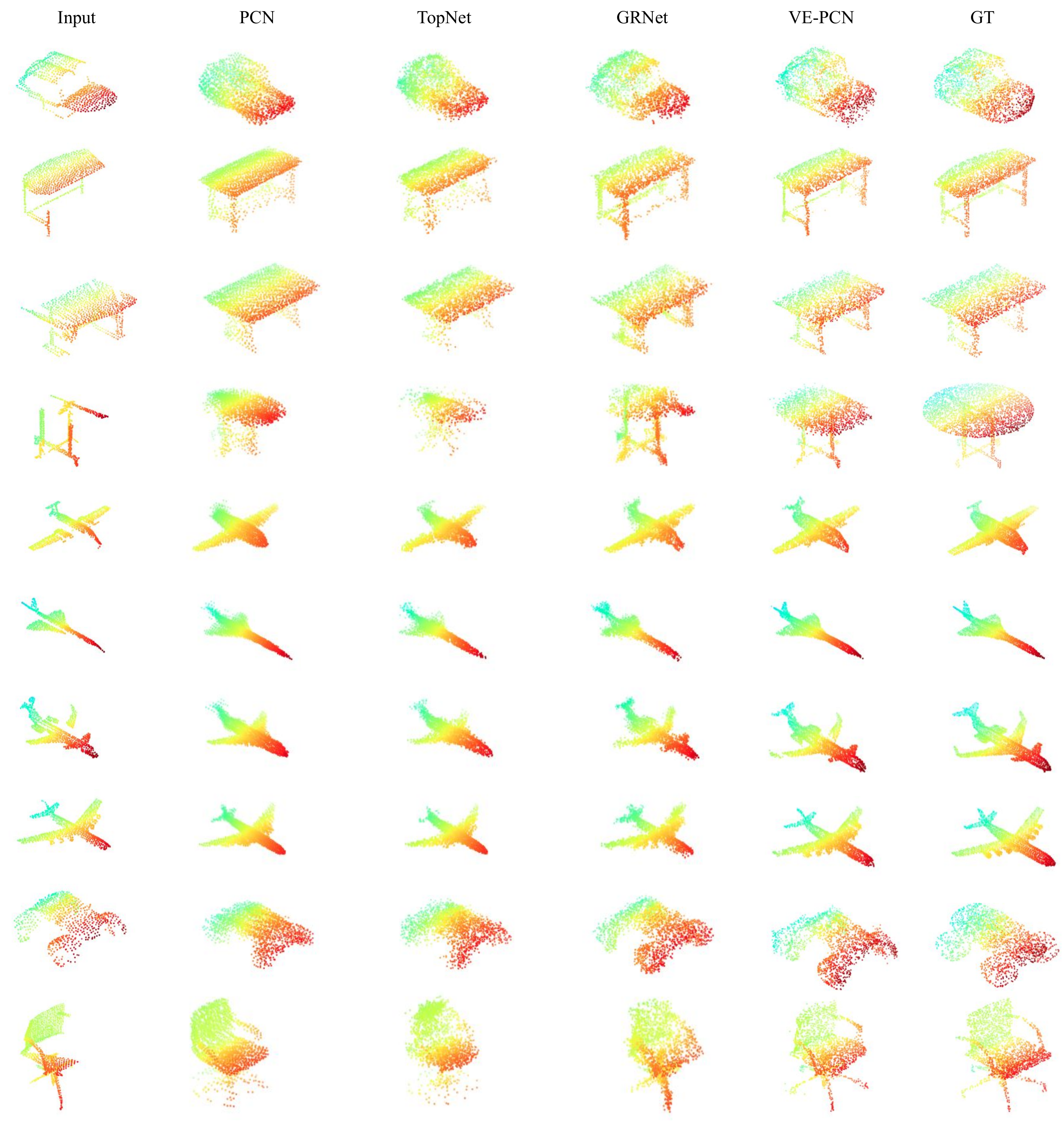}
  \caption{Qualitative comparisons on the Completion3D dataset (1/2).}
  \label{topnet_supp_1}
\end{figure*}

\begin{figure*}
    \centering
  \includegraphics[width=\linewidth]{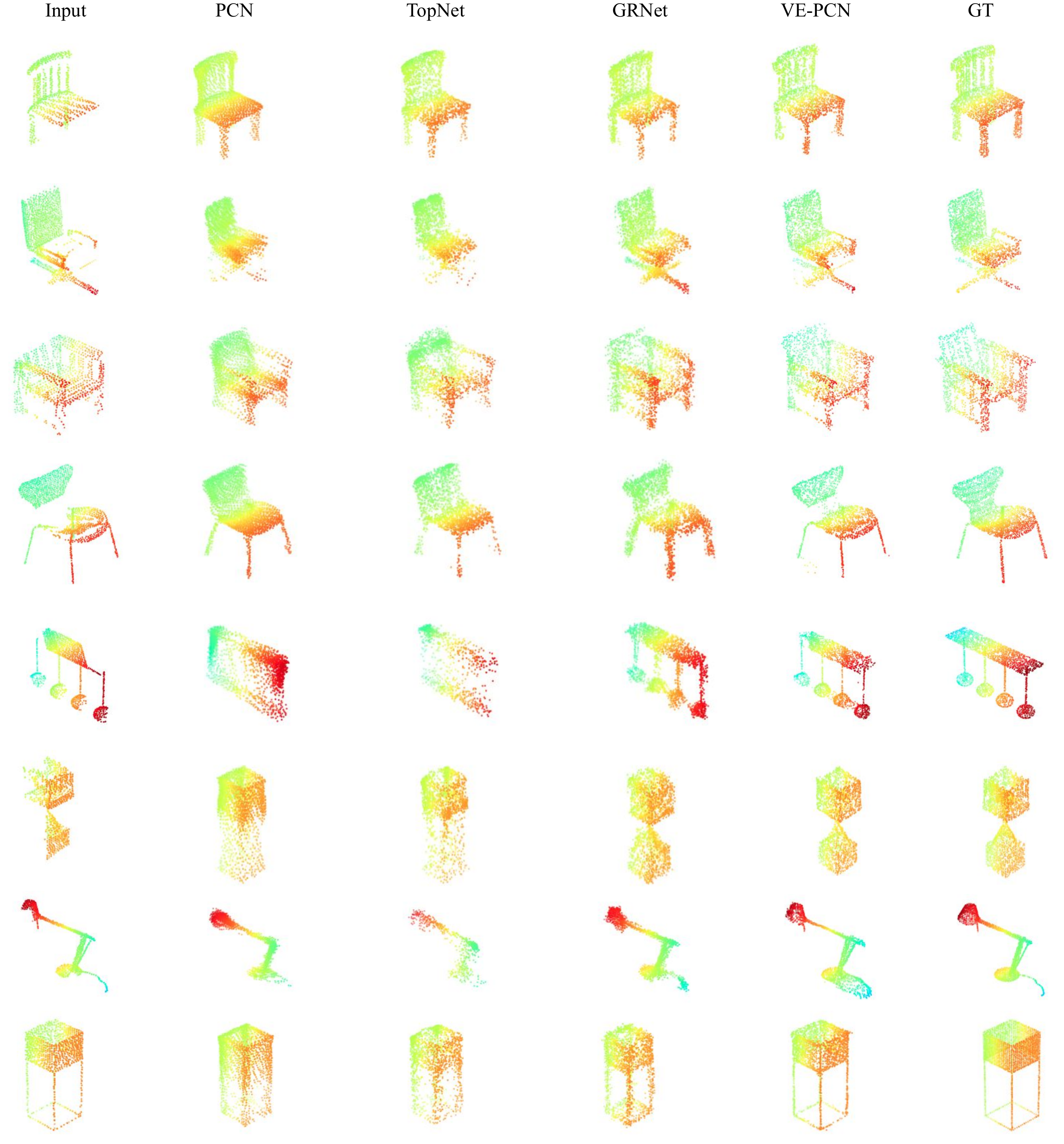}
  \caption{Qualitative comparisons the Completion3D dataset (2/2).}
  \label{topnet_supp_2}
\end{figure*}

More qualitative results on the validation data are shown in Figures~\ref{topnet_supp_1} and~\ref{topnet_supp_2}.


\subsection{More Results on our Dataset}
\begin{figure*}
    \centering
  \includegraphics[width=\linewidth]{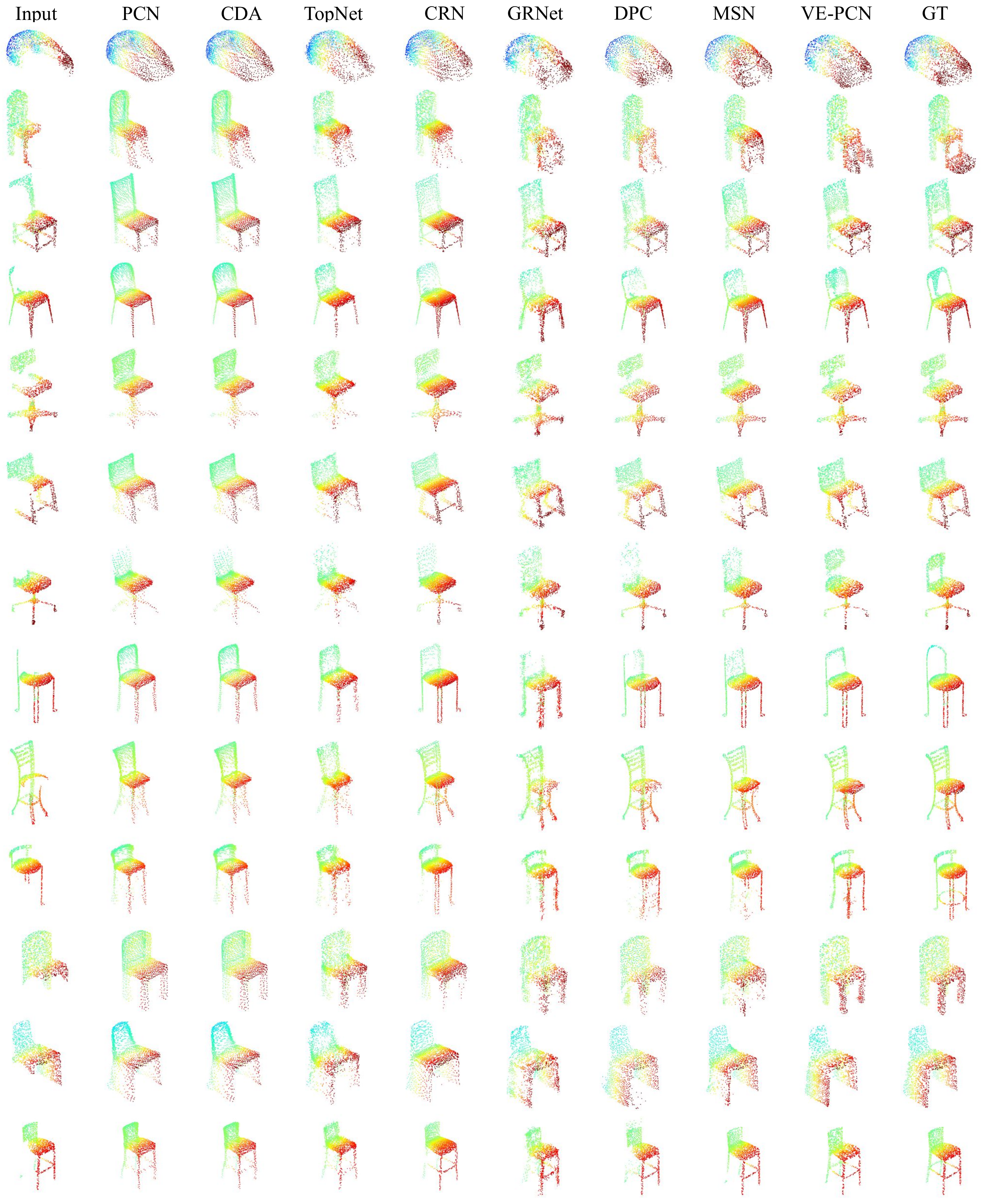}
  \caption{Qualitative comparisons on seen categories of our dataset (1/2).}
  \label{quantative_seen_supp_1}
\end{figure*}

\begin{figure*}
    \centering
  \includegraphics[width=\linewidth]{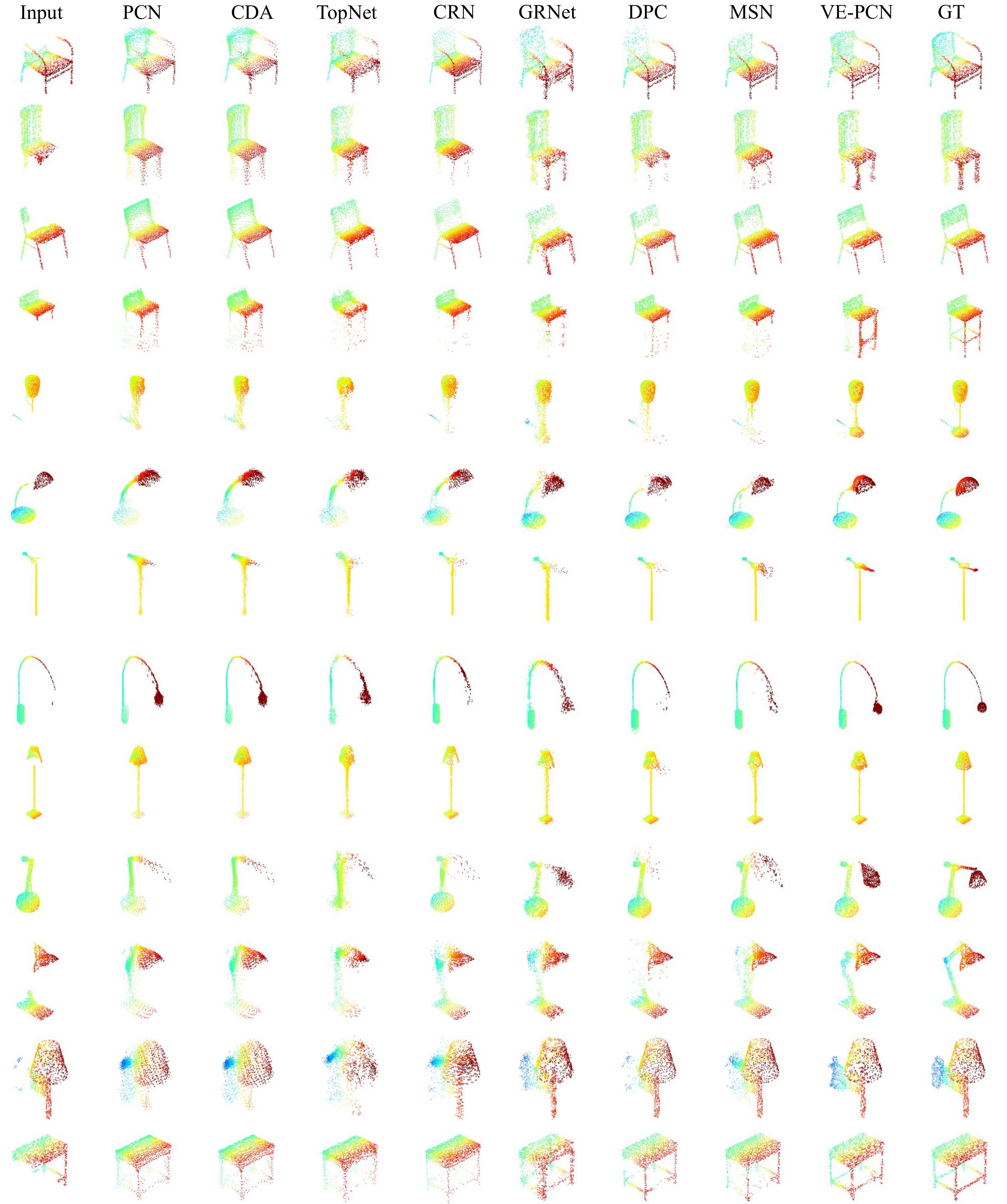}
  \caption{Qualitative comparisons on seen categories of our dataset (2/2).}
  \label{quantative_seen_supp_2}
\end{figure*}

\begin{figure*}
    \centering
  \includegraphics[width=\linewidth]{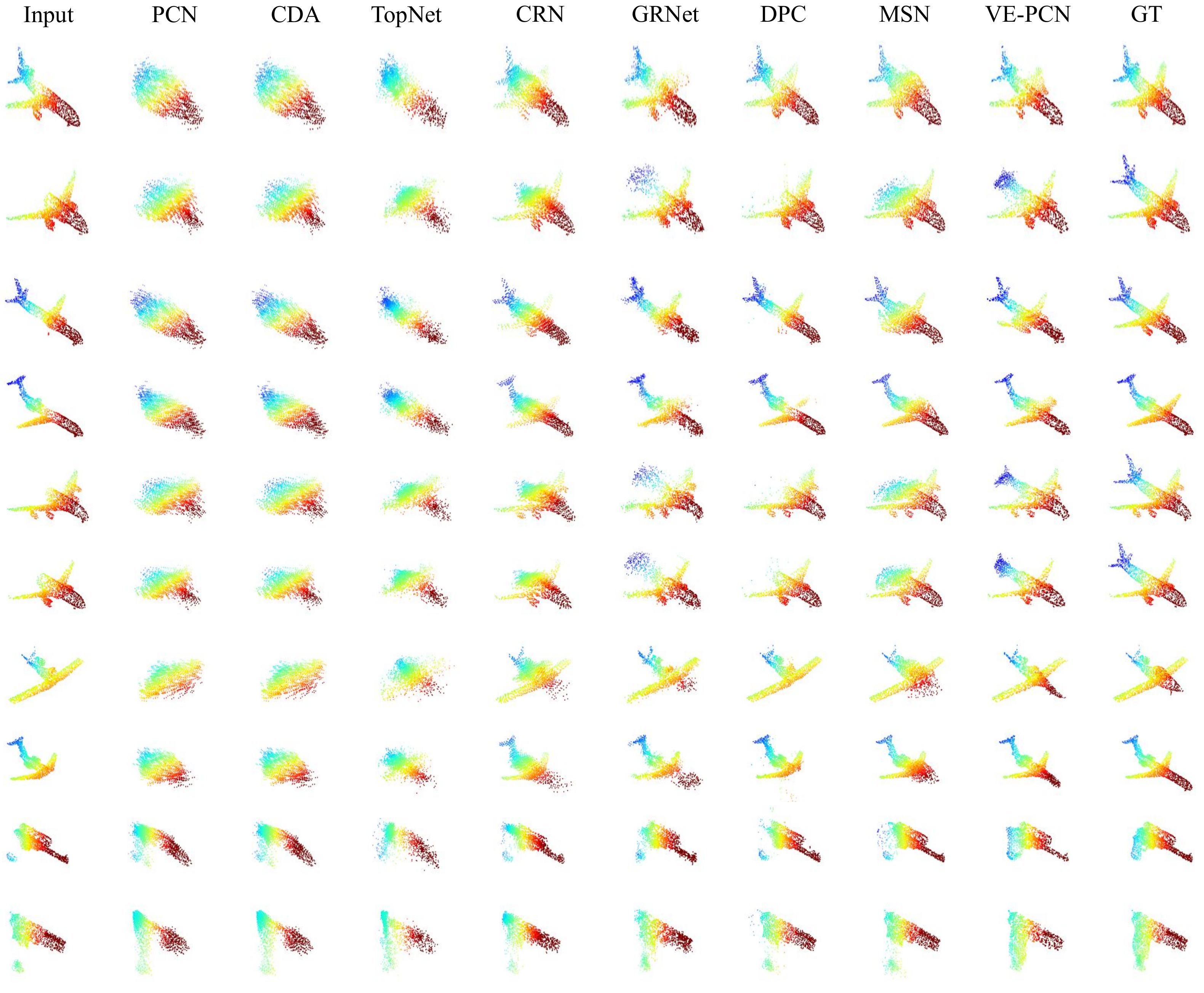}
  \caption{Qualitative comparisons on unseen categories of our dataset.}
  \label{qualitative_seen_supp_unseen}
\end{figure*}

Table~\ref{quantative_seen_fpd} shows the FPD evaluations from various methods. 
More qualitative results on seen categories of our dataset are shown in Figures~\ref{quantative_seen_supp_1} and~\ref{quantative_seen_supp_2}. 
More qualitative results on unseen categories are shown in Figure~\ref{qualitative_seen_supp_unseen}.

\begin{figure*}
    \centering
  \includegraphics[width=0.93\linewidth]{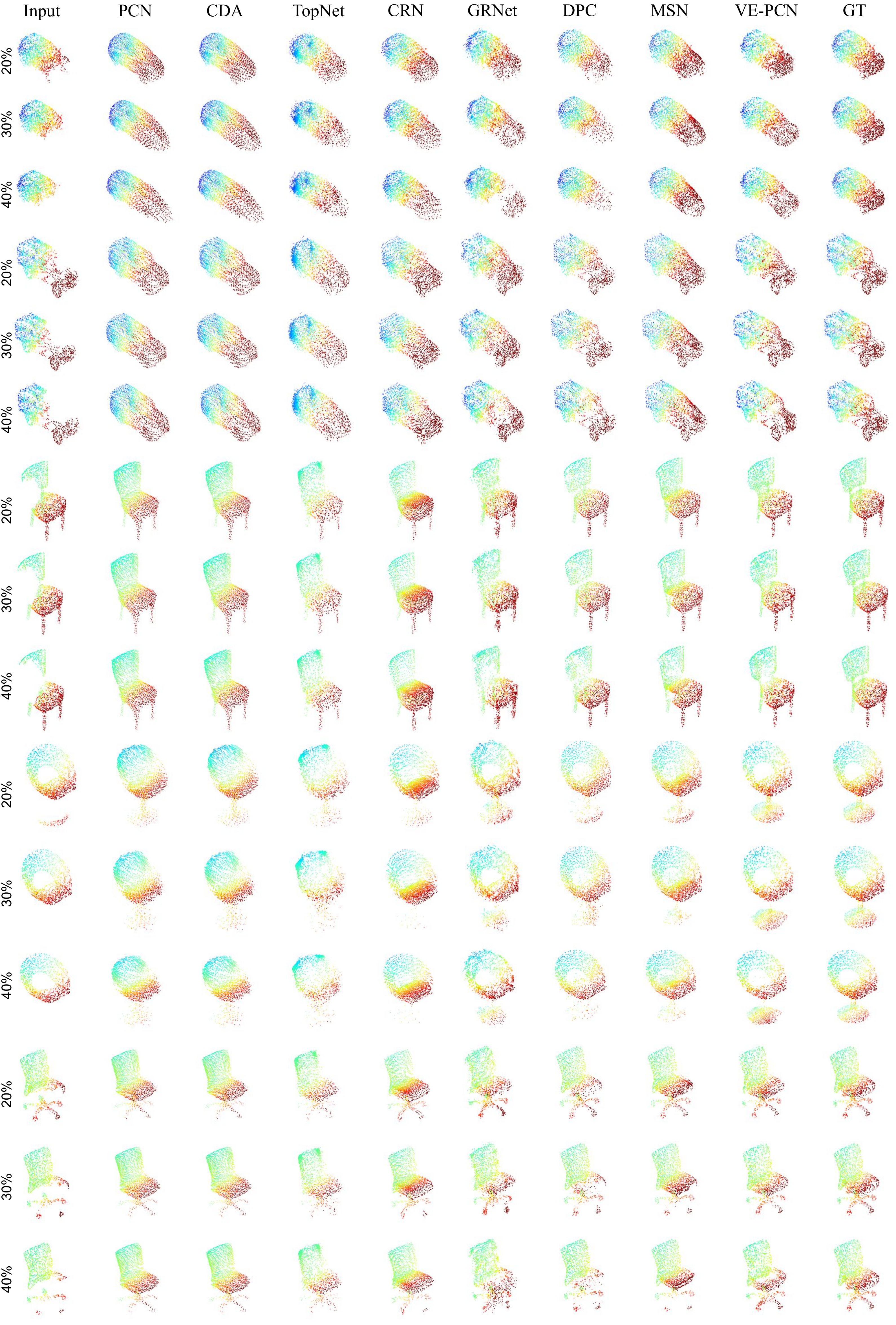}
  \vspace{-3mm}
  \caption{Qualitative comparisons on different occlusion ratios (1/2).}
  \label{different_ratios}
\end{figure*}

\begin{figure*}
    \centering
  \includegraphics[width=0.93\linewidth]{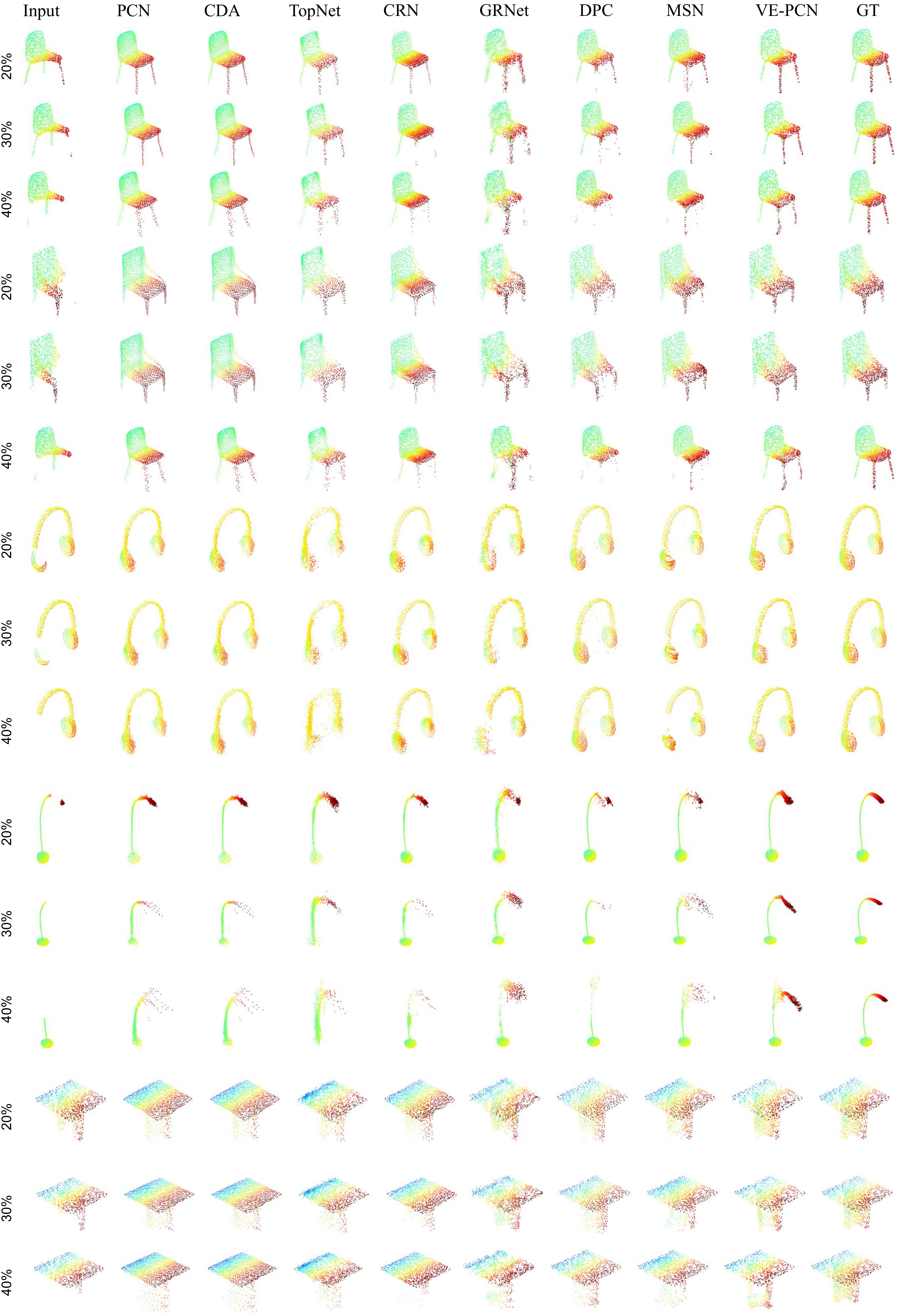}
  \caption{Qualitative comparisons on different occlusion ratios (2/2).}
  \label{different_ratios_2}
\end{figure*}

To further test the robustness of different models, we occlude the partial input with different occlusion ratios $p$ that ranges from $20\%$ to $40\%$. 
We directly adopt the models trained on seen categories for testing.
The quantitative results are shown in Table~\ref{quantitative_occlusion}.
More qualitative results are shown in Figures~\ref{different_ratios} and ~\ref{different_ratios_2}.

\subsection{More Results on the KITTI Dataset}
\begin{figure}
    \centering
  \includegraphics[width=\linewidth]{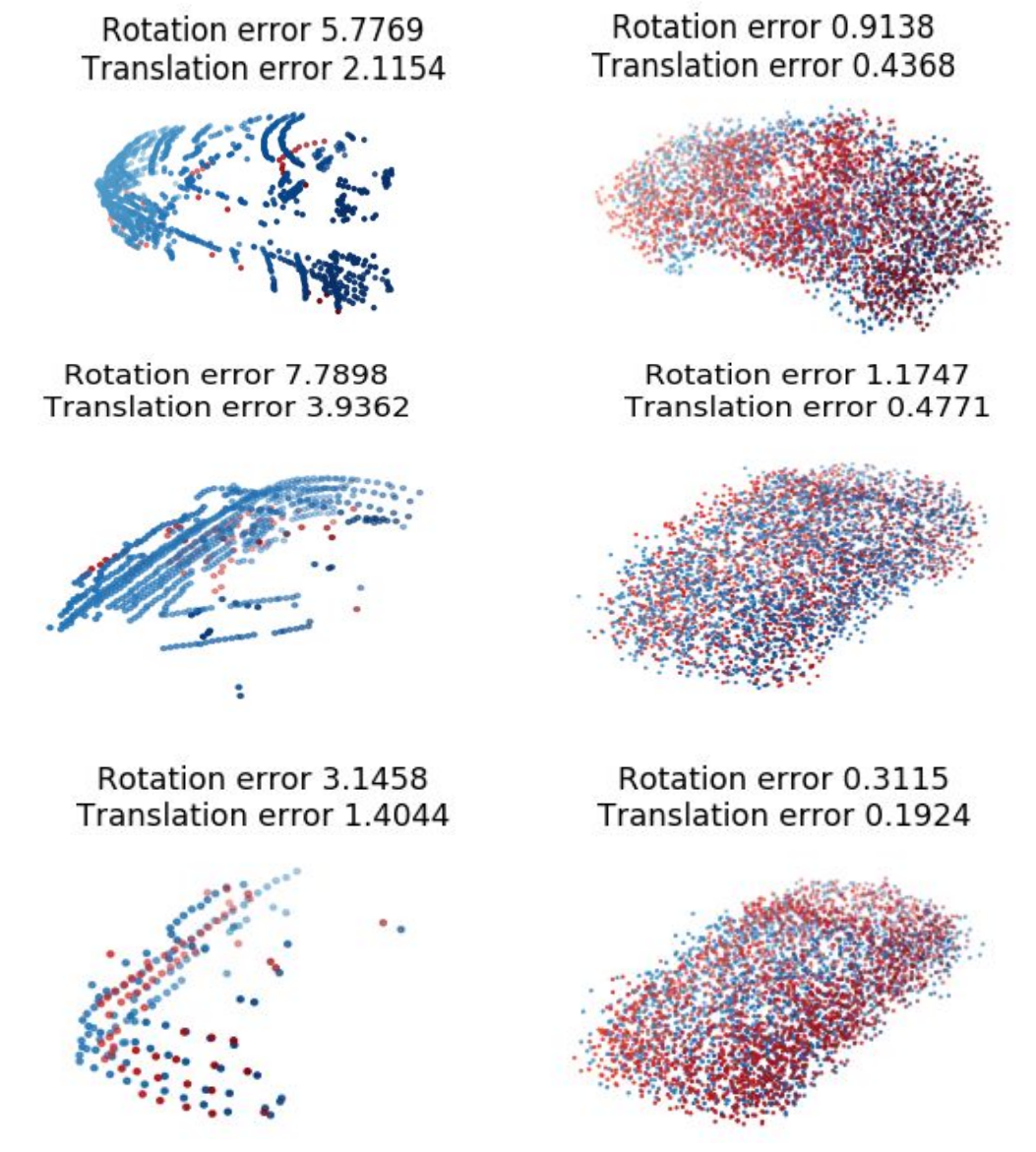}
  \caption{Completion results on KITTI.}
  \label{kitti_supp}
\end{figure}

Figure~\ref{kitti_supp} shows the completion results on the KITTI dataset. We evaluate the performances by calculating the registration errors following PCN~\cite{yuan2018pcn}. 

\subsection{More Results on the Edge Generation}
\begin{figure}
    \centering
  \includegraphics[width=\linewidth]{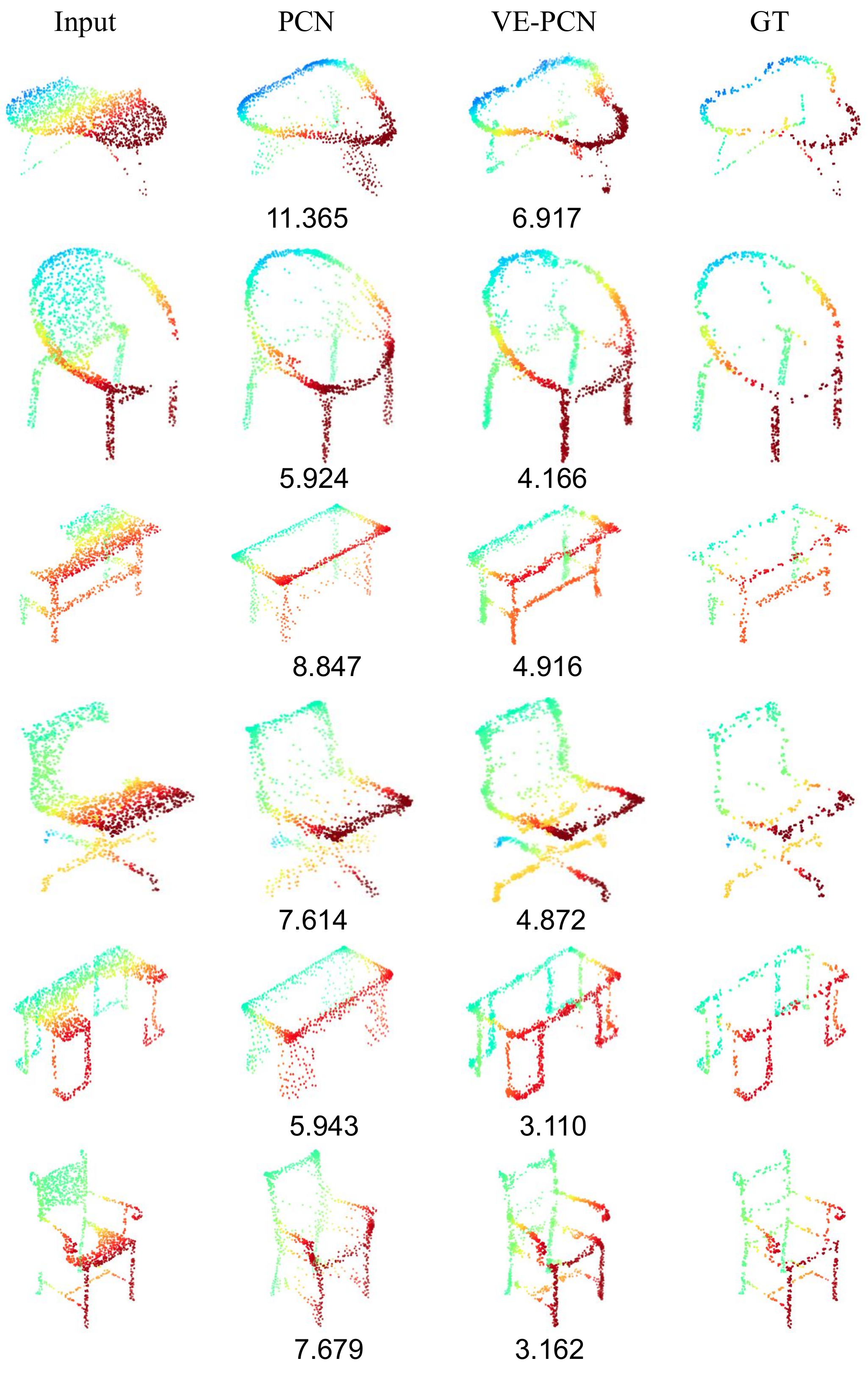}
  \caption{Edge generation results. Bottom values are the CD errors per point ($10^{-4}$).}
  \label{edges_supp}
\end{figure}

We show more experimental results on point cloud edge generation in Figure~\ref{edges_supp}. 

\section{Comparisons to SK-PCN~\cite{NEURIPS2020_ba036d22}}
\begin{figure*}
\centering
  \includegraphics[width=0.93\linewidth]{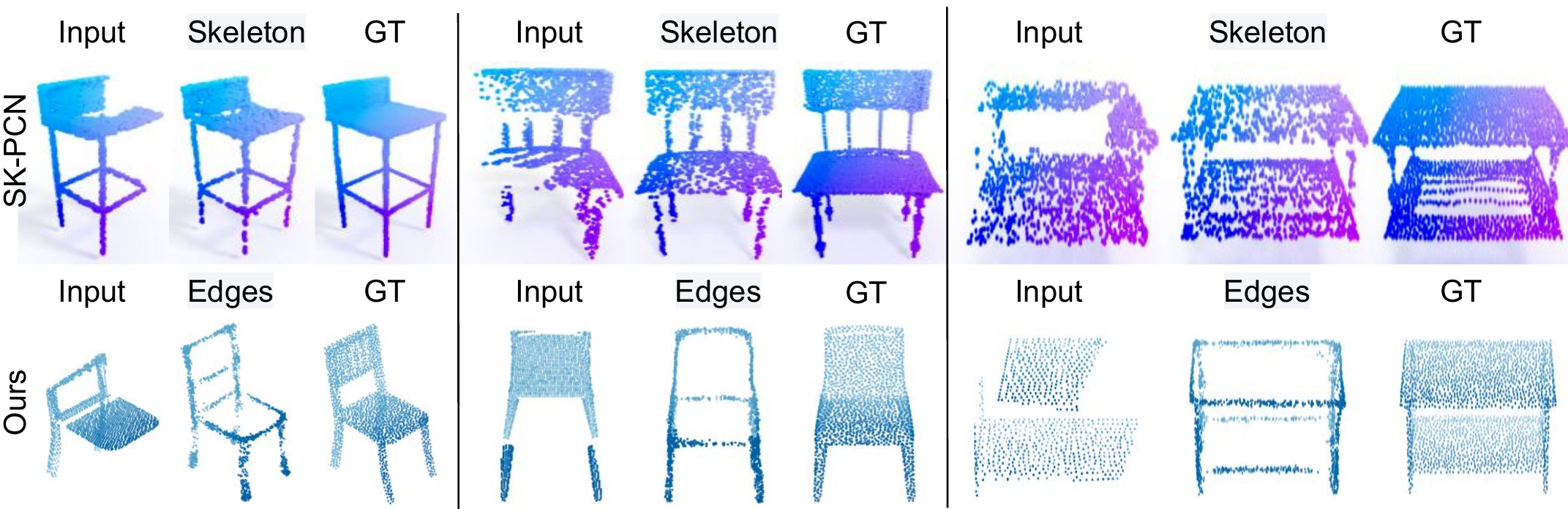}
  \caption{Edge (ours) vs. mes-skeleton (SK-PCN) generation.}
  \vspace{-0.6cm}
  \label{skpcn}
\end{figure*}

SK-PCN proposes a similar thought that adopts skeleton generations to help the shape completion. 
However, our edges are different from their meso-skeleton. The differences are shown in Figure~\ref{skpcn} (The results of SK-PCN are Fig. 9 of their paper).
Their meso-skeleton focus on the overall shapes.
In contrast, our edges focus on high frequency components (\emph{e.g.} thin structures), which are difficult to generate in existing methods. This can be evidenced from the limitation cases in Fig. 10 of SK-PCN paper.
Moreover, SK-PCN generates the complete points by learning displacements from skeletal points with a local adjustment strategy.
In contrast, we synthesize the complete points by injecting the edge features into the completion decoder with a voxelization strategy. 
This voxel structure further enables our point generation to be confined within well-defined spaces of the grid cells, and thus eliminates the generation of spurious points that are commonly seen in other methods.



\clearpage

{\small
\bibliographystyle{ieee_fullname}
\bibliography{egbib}
}

\end{document}

%% file: tables/quantative_topnet.tex
\begin{table*}[!htbp]
\centering
\resizebox{0.85\textwidth}{!}{
\small
\begin{tabular}{l|c| c |c  |c  | c|  c|  c |c|c }
\hline
\multirow{2}{*}{Methods} & \multicolumn{9}{c}{Mean Chamfer Distance per point ($10^{-4}$)} \\
\cline{2-10}
{}& Avg & Airplane & Cabinet & Car & Chair & Lamp & Sofa & Table & Vessel \\
\hline\hline
FoldingNet\cite{yang2018foldingnet} &19.07 &12.83 &23.01 &14.88 &25.69 &21.79 &21.31 &20.71 &11.51 \\
PCN\cite{yuan2018pcn} &18.22 &9.79 &22.70 &12.43 &25.14 &22.72 &20.26 &20.27 &11.73 \\
PointSetVoting\cite{zhang2021point} &18.18 &6.88 &21.18 &15.78 &22.54 &18.78 &28.39 &19.96 &11.16 \\
AtlasNet\cite{groueix2018} &17.77 &10.36 &23.40 &13.40 &24.16 &20.24 &20.82 &17.52 &11.62 \\
TopNet~\cite{topnet2019} &14.25 &7.32 &18.77 &12.88 &19.82 &14.60 &16.29 &14.89 &8.82 \\
SoftPoolNet\cite{wang2020softpoolnet} &11.90 &4.89 &18.86 &10.17 &15.22 &12.34 &14.87 &11.84 &6.48 \\
SA-Net\cite{wen2020point} &11.22 &5.27 &14.45 &7.78 &13.67 &13.53 &14.22 &11.75 &8.84 \\
GRNet\cite{xie2020grnet} &10.64 &6.13 &16.90 &8.27 &12.23 &10.22 &14.93 &10.08 &5.86 \\
PMP-Net\cite{wen2021pmp} &9.23 &3.99 &14.70 &8.55 &10.21 &9.27 &12.43 &8.51 &5.77 \\
CRN\cite{Wang_2020_CVPR} &9.21 &3.38 &13.17 &8.31 &10.62 &10.00 &12.86 &9.16 &5.80 \\
SCRN\cite{wang2020self} &9.13 &\textbf{3.35} &12.81 &\textbf{7.78} &9.88 &10.12 &12.95 &9.77 &6.10 \\
VE-PCN &\textbf{8.10} &3.83 &\textbf{12.74} &7.86 &\textbf{8.66} &\textbf{7.24} &\textbf{11.47} &\textbf{7.88} &\textbf{4.75} \\
\hline
\end{tabular}
}
\vspace{-3mm}
\caption{Quantitative comparison for point cloud completion on eight categories objects of Completion3D benchmark.}
\vspace{-2mm}
\label{quantative_topnet}
\end{table*}

%% file: tables/quantative_seen.tex
\begin{table*}[!htbp]
\centering
\resizebox{1\textwidth}{!}{
\small
\begin{tabular}{l|c| c |c  |c  | c|  c|  c|  c|  c}
\hline
\multirow{2}{*}{Categories} & \multicolumn{8}{c}{Mean Chamfer Distance per point ($10^{-4}$)} \\
\cline{2-10}
{}& PCN~\cite{yuan2018pcn} & PCN-FC~\cite{yuan2018pcn} & CDA~\cite{lim2019convolutional} & TopNet~\cite{topnet2019} & CRN~\cite{Wang_2020_CVPR} & GRNet~\cite{xie2020grnet} & DPC~\cite{zhang2020detail} & MSN~\cite{liu2020morphing} & VE-PCN\\
\hline\hline
Bag &11.609 &11.722 &15.854 &13.571 &6.253 &6.048 &4.543 &6.118 &\textbf{3.466}  \\
Cap &9.451 &11.089 &14.437 &13.191 &6.709 &6.334 &3.218 &4.918 &\textbf{3.098}  \\
Car &7.254 &7.667 &8.344 &8.419 &4.776 &5.704 &3.714 &8.214 &\textbf{3.480}  \\
Chair &4.675 &5.165 &6.446 &6.026 &3.116 &4.286 &2.811 &2.666 &\textbf{2.476}  \\
Earphone &10.821 &10.437 &14.026 &11.921 &6.117 &5.321 &4.991 &8.309 &\textbf{3.239}  \\
Guitar &0.996 &1.134 &1.289 &1.403 &0.690 &1.309 &0.869 &\textbf{0.641} &0.750  \\
Lamp &9.421 &10.592 &12.225 &10.619 &5.652 &4.443 &5.809 &5.618 &\textbf{3.226}  \\
Laptop &3.236 &3.464 &3.653 &3.941 &2.534 &3.745 &1.854 &\textbf{1.639} &2.197 \\
Motorbike &6.005 &6.398 &6.706 &7.033 &3.434 &4.237 &4.081 &4.692 &\textbf{2.717} \\
Mug &9.522 &10.788 &11.877 &11.706 &6.731 &8.022 &6.802 &5.407 &\textbf{4.938} \\
Skateboard &3.504 &3.858 &4.464 &4.321 &2.288 &2.835 &1.804 &2.431 &\textbf{1.780} \\
Table &5.990 &6.789 &8.010 &6.973 &3.805 &4.325 &3.946 &3.199 &\textbf{2.798} \\
\hline
Average &5.771 &6.420 &7.577 &6.889 &3.697 &4.261 &3.570 &3.550 &\textbf{2.669} \\
\hline
\end{tabular}
}
\vspace{-3mm}
\caption{Quantitative comparison for point cloud completion on 12 seen categories of our dataset.}
\vspace{-3mm}
\label{quantative_seen}
\end{table*}

%% file: tables/quantative_diff_resolutions.tex
\begin{table*}[!htbp]
\centering
\resizebox{1\textwidth}{!}{
\small
\begin{tabular}{l|c| c |c  |c  | c|  c|  c|  c|  c}
\hline
\multirow{2}{*}{Resolutions} & \multicolumn{8}{c}{Mean Chamfer Distance per point ($10^{-4}$)} \\
\cline{2-10}
{}& PCN~\cite{yuan2018pcn} & PCN-FC~\cite{yuan2018pcn} & CDA~\cite{lim2019convolutional} & TopNet~\cite{topnet2019} & CRN~\cite{Wang_2020_CVPR} & GRNet~\cite{xie2020grnet} & DPC~\cite{zhang2020detail} & MSN~\cite{liu2020morphing} & VE-PCN\\
\hline\hline
2048 &5.771 &6.420 &7.577 &6.889 &3.697 &4.261 &3.570 &3.550 &\textbf{2.669} \\
4096 &3.821 &4.472 &6.721 &4.588 &2.671 &2.956 &2.943 &3.409 &\textbf{2.275}  \\
8192 &3.264 &3.795 &6.100 &4.319 &2.342 &2.423 &2.920 &3.808 &\textbf{1.880}  \\
16384 &2.864 &3.251 &5.934 &3.513 &1.953 &2.011 &2.462 &3.588 &\textbf{1.620} \\
\hline
\end{tabular}
}
\vspace{-3mm}
\caption{Quantitative comparison of point cloud completion with different resolutions of our dataset.
}
\vspace{-3mm}
\label{quantative_diff_resolutions}
\end{table*}

%% file: tables/quantative_unseen.tex
\begin{table*}[!htbp]
\centering
\resizebox{1\textwidth}{!}{
\small
\begin{tabular}{l|c| c |c  |c  | c|  c|  c|  c| c}
\hline
\multirow{2}{*}{Categories} & \multicolumn{8}{c}{Mean Chamfer Distance per point ($10^{-4}$)} \\
\cline{2-10}
{}& PCN~\cite{yuan2018pcn} & PCN-FC~\cite{yuan2018pcn} & CDA~\cite{lim2019convolutional} & TopNet~\cite{topnet2019} & CRN~\cite{Wang_2020_CVPR} & GRNet~\cite{xie2020grnet} & DPC~\cite{zhang2020detail} & MSN~\cite{liu2020morphing} &VE-PCN\\
\hline\hline
Airplane &26.418  &28.518 &33.550 &17.675 &11.047 &4.526 &10.301 &9.906 &\textbf{3.382} \\
Knife &2.998 &3.110 &5.235 &3.495 &2.015 &1.700 &3.518 &2.604 &\textbf{1.603}  \\
Pistol &12.168 &11.131 &27.701 &11.038 &6.428 &\textbf{3.482} &6.505 &5.524 &4.884  \\
Rocket &10.035 &10.853 &16.029 &8.075 &4.750 &1.999 &5.378 &4.992 &\textbf{1.574}  \\
\hline
Average &20.764 &22.208 &27.821 &14.440 &8.948 &3.892 &8.690 &8.154 &\textbf{3.176}  \\
\hline
\end{tabular}
}
\vspace{-3mm}
\caption{Quantitative comparison for point cloud completion on unseen categories of our dataset.}
\vspace{-2mm}
\label{quantative_unseen}
\end{table*}

%% file: tables/kitti_quantitative.tex
\begin{table*}[!htbp]
\centering
\resizebox{1\textwidth}{!}{
\small
\begin{tabular}{l|c| c |c  |c  | c|  c|  c|  c|  c}
\hline
Methods & PCN~\cite{yuan2018pcn} & PCN-FC~\cite{yuan2018pcn} & CDA~\cite{lim2019convolutional} &TopNet~\cite{topnet2019}  & CRN~\cite{Wang_2020_CVPR} & GRNet~\cite{xie2020grnet} & DPC~\cite{zhang2020detail} & MSN~\cite{liu2020morphing} & VE-PCN\\
\hline
Fidelity &0.0436 &0.0407 &0.0428 &0.0438 &0.0337 &0.0298 &0.0347 &0.0345 &\textbf{0.0258} \\
\hline
\end{tabular}
}
\vspace{-3mm}
\caption{Fidelity evaluations that measure the average distance between each input point to its nearest neighbor in the output (lower is better) on the KITTI dataset.}
\vspace{-3mm}
\label{kitti_quantitative}
\end{table*}

%% file: tables/ablation.tex
\begin{table}[!htbp]
\centering
\small
\vspace{-3mm}
\begin{tabular}{l|c| c| c |c  |c }
\hline
ID&MGT & DG & EG & PE &CD\\
\hline
1& \cmark &\xmark &\cmark &\xmark &\textbf{2.669} \\
2& \cmark &\xmark &\xmark &\cmark &3.586 \\
3& \cmark &\xmark &\xmark &\xmark &3.600 \\
4& \xmark &\cmark &\xmark &\xmark &4.768 \\
5& \xmark &\xmark &\xmark &\xmark &5.792 \\
\hline
\end{tabular}
\vspace{-3mm}
\caption{Ablation studies on our dataset. Results are represented by mean CD per point (10$^{-4}$).}
\vspace{-6mm}
\label{ablation}
\end{table}

%% file: tables_supp/model_size.tex
\begin{table*}[]
\centering
\resizebox{0.95\textwidth}{!}{
\begin{tabular}{c| c| c| c|c| c|c|c|c|c}
\hline
 & PCN~\cite{yuan2018pcn} & PCN-FC~\cite{yuan2018pcn} & CDA~\cite{lim2019convolutional} &TopNet~\cite{topnet2019}  & CRN~\cite{Wang_2020_CVPR} & GRNet~\cite{xie2020grnet} & DPC~\cite{zhang2020detail} & MSN~\cite{liu2020morphing} & VE-PCN\\
\hline
Para. (M) &6.85 &53.2 &51.85 &9.96 &5.14 &76.71 &6.66 &30.32 &35.00 \\
Time (ms) &57.5 &21.4 &614.9 &63.1 &61.3 &124.3 &331.3 &346.7 &450.1 \\
\hline
\end{tabular}
}
\caption{Space and time comparisons of different methods.}
\label{model_size}
\end{table*}

%% file: tables_supp/ablation_supp.tex
\begin{table}[!htbp]
\centering
\small
\begin{tabular}{c| c}
\hline
Methods & CD\\
\hline
full pipeline &\textbf{2.669} \\
$\lambda_{2}=0$ &2.886 \\
$\lambda_{1}=0$ &3.408 \\
$\lambda_{4}=0$ &2.843 \\
$\lambda_{3}=0$ &2.842 \\
\hline
\end{tabular}
\caption{Ablation studies on the different losses. Results are obtained by evaluating mean CD per point (10$^{-4}$) on our dataset.}
\label{ablation_supp}
\end{table}

%% file: tables/pcn_results.tex
\begin{table*}[!htbp]
\centering
\vspace{-2mm}
\resizebox{0.8\textwidth}{!}{
\small
\begin{tabular}{ c|c| c |c  |c  | c|  c|  c |c|c }
\hline
\multirow{2}{*}{Methods} & \multicolumn{9}{c}{Mean Chamfer Distance (CD) per point ($10^{-3}$)} \\
\cline{2-10}
& Average & Plane & Cabinet & Car & Chair & Lamp & Sofa & Table & Vessel \\
\hline\hline
FoldingNet~\cite{yang2018foldingnet} & 14.31 & 9.49 & 15.80 & 12.61 & 15.55 & 16.41 & 15.97 & 13.65 & 14.99 \\
TopNet~\cite{topnet2019} &12.15 &7.61 &13.31 &10.90 &13.82 &14.44 &14.78 &11.22 &11.12 \\
AtlasNet~\cite{groueix2018} & 10.85 & 6.37 & 11.94 & 10.10 & 12.06 & 12.37 & 12.99 & 10.33 & 10.61 \\
PCN~\cite{yuan2018pcn} &9.64 &5.50 &10.63 &8.70 &11.00 &11.34 &11.68 &8.59 &9.67\\
GRNet~\cite{xie2020grnet} &8.83 &6.45 &10.37 &9.45 &9.41 &7.96 &10.51 &8.44 &8.04 \\
CRN~\cite{Wang_2020_CVPR} &8.51 &\textbf{4.79} &9.97 &\textbf{8.31} &9.49 &8.94 &10.69 &7.81 &8.05 \\
PMP-Net~\cite{wen2021pmp} &8.66 &5.50 &11.10 &9.62 &9.47 &\textbf{6.89} &10.74 &8.77 &\textbf{7.19} \\
VE-PCN &\textbf{8.32} &4.80 &\textbf{9.85} &9.26 &\textbf{8.90} &8.68 &\textbf{9.83} &\textbf{7.30} &7.93 \\
\hline
\end{tabular}
}
\caption{Quantitative results on the PCN dataset.
}
\label{pcn_results}
\end{table*}

%% file: tables/quantative_seen_fpd.tex
\begin{table*}[!htbp]
\centering
\resizebox{1\textwidth}{!}{
\small
\begin{tabular}{l|c|c |c |c | c|  c|  c|  c|  c}
\hline
{}& PCN~\cite{yuan2018pcn} & PCN-FC~\cite{yuan2018pcn} & CDA~\cite{lim2019convolutional} & TopNet~\cite{topnet2019} & CRN~\cite{Wang_2020_CVPR}  & GRNet~\cite{xie2020grnet} & DPC~\cite{zhang2020detail}  & MSN~\cite{liu2020morphing}& VE-PCN\\
\hline
FPD &5.584 &6.634 &9.142 &7.7536 &3.054 &6.513 &8.347 &3.904 &\textbf{1.882}  \\
\hline
\end{tabular}
}
\caption{FPD comparisons on different methods. The lower, the better.}
\label{quantative_seen_fpd}
\end{table*}

%% file: tables_supp/quantitative_occlusion.tex
\begin{table*}[!htbp]
\centering
\resizebox{0.96\textwidth}{!}{
\small
\begin{tabular}{l|c| c |c  |c  | c|  c|  c|  c|  c}
\hline
\multirow{2}{*}{Ratios} & \multicolumn{8}{c}{Mean Chamfer Distance per point ($10^{-4}$)} \\
\cline{2-10}
{}& PCN~\cite{yuan2018pcn} & PCN-FC~\cite{yuan2018pcn} & CDA~\cite{lim2019convolutional} & TopNet~\cite{topnet2019} & CRN~\cite{Wang_2020_CVPR} & GRNet~\cite{xie2020grnet} & DPC~\cite{zhang2020detail} & MSN~\cite{liu2020morphing} & VE-PCN\\
\hline\hline
20\% &5.642 &6.230 &7.485 &6.732 &3.554 &4.016 &2.975 &3.213 &\textbf{2.565} \\
30\% &5.991 &6.704 &7.771 &7.245 &3.932 &4.646 &4.681 &4.084 &\textbf{3.055} \\
40\% &7.066 &7.934 &8.741 &8.622 &\textbf{5.094} &10.746 &8.210 &6.546 &5.129 \\
\hline
\end{tabular}
}
\caption{Quantitative comparison of occluded point clouds under different occlusion rates.}
\label{quantitative_occlusion}
\end{table*}